\pdfoutput=1

\documentclass[10pt,twocolumn,letterpaper]{article}

\usepackage{cvpr}
\usepackage{times}
\usepackage{epsfig}
\usepackage{graphicx}
\usepackage{amsmath}
\usepackage{amssymb}
\usepackage{algorithm}
\usepackage{algorithmic}
\usepackage{rotating}
\usepackage[T1]{fontenc}

\usepackage[breaklinks=true,bookmarks=false]{hyperref}

\cvprfinalcopy 

\def\cvprPaperID{1967} 

\begin{document}
\title{Learning the Matching Function}

\author{\v Lubor Ladick\'y\\
ETH Z\"urich, Switzerland\\
{\tt\small lubor.ladicky@inf.ethz.ch}
\and
 Christian H\"ane\\
ETH Z\"urich, Switzerland\\
{\tt\small chaene@inf.ethz.ch}
\and
Marc Pollefeys\\
ETH Z\"urich, Switzerland\\
{\tt\small marc.pollefeys@inf.ethz.ch}
}

\maketitle

\newcommand{\comment}[1]{}

\begin{abstract}
The matching function for the problem of stereo reconstruction or optical flow has been traditionally designed as a function of the distance between the features describing matched pixels. This approach works under assumption, that the appearance of pixels in two stereo cameras or in two consecutive video frames does not change dramatically. However, this might not be the case, if we try to match pixels over a large interval of time.

In this paper we propose a method, which learns the matching function, that automatically finds the space of allowed changes in visual appearance, such as due to the motion blur, chromatic distortions, different colour calibration or seasonal changes. Furthermore, it automatically learns the importance of matching scores of contextual features at different relative locations and scales. Proposed classifier gives reliable estimations of pixel disparities already without any form of regularization.

We evaluated our method on two standard problems - stereo matching on KITTI outdoor dataset, optical flow on Sintel data set, and on newly introduced TimeLapse change detection dataset.
Our algorithm obtained very promising results comparable to the state-of-the-art.
\end{abstract}

\section{Introduction}

The matching problems, such as dense stereo reconstruction or optical flow, typically uses as an underlying similarity measure between pixels the distance between patch-features surrounding the pixels. In many scenarios, for examples when high quality rectified images are available, this approach is sufficient to find reliable pixel correspondences between images. To get smooth results often a regularization in the discrete or continuous CRF framework is utilized. The research in this field has been focused on finding the most robust features, designing the best distance measure, choosing the right form of regularization and developing fast and reliable optimization methods for the corresponding optimization problem. An exhaustive overview about matching algorithms can be found in~\cite{Scharstein02} and a comparison of different dissimilarity measures can be found in~\cite{Hirschmuller07}. The
proposed methods are fully generative and do not require any form of discriminative training, except of the weight of regularization, which can be hand-tuned
without much effort.

The main problem of such approaches occurs if the scene contains large texture-less regions. In that case each pixel matches with any other pixel and the result is determined solely by the regularization, which typically biases the solution towards a constant labelling. This causes severe problems for large planar surfaces, such as walls or a ground plane. For many datasets, this is typically resolved using special dataset-dependent priors; for example for the KITTI dataset~\cite{Geiger2012CVPR} using the known height of the camera above the ground plane. To get state-of-the-art results it is necessary to over-engineer the method for each particular dataset. For an optical flow problem, this includes heuristics, solving the problem in a coarse-to-fine fashion, constraining the allowed range of disparities or flows based on matched sparse features~\cite{Brox11} or recursive deep matching in a spatial pyramid~\cite{Weinzaepfel13}. Such approaches lead to a very large boost of performance quantitatively, however they often miss thin structures, due to the inability of the coarse-to-fine procedure to recover from incorrect label in the higher level of the pyramid.

To find the right matching function, the researchers typically focus on search for the most robust feature. Such solutions are often sufficient, because the appearance of pixels in two stereo cameras or in two consecutive video frames does not change dramatically. However, this is not the case, when we try to decide, what has changed in the scene within a large interval of time (such as half a year). Depending on the application, we might be interested in what buildings have been built or if there was the ship present in the port. The importance of temporal change detection has been recognized~\cite{Taneja13} for projects such as Google Street View with an ultimate goal of building up-to-date 3D models of cities, while minimizing the costs of the updates. The problem pixel-wise temporal change detection can also be cast as a matching problem; transformation to which the matching function has to be invariant significantly increases. For example for city scenes the matching function has to allow for visual changes not only due to different lighting conditions, but also due to seasonal changes, such as the change of appearance of leaves of the tree during the year or presence of the snow on the ground.

In this paper we propose a method, which learns a matching function, that automatically finds the space of allowed changes in visual appearance, such as due to the motion blur, chromatic aberrations, different colour calibration or seasonal changes. We evaluated our method on pixel-wise temporal change detection, for which we introduce a new TimeLapse change detection dataset. The dataset contains pairs of images taken at different time of the year from the same spot, and the human-labelled ground truth, determining what has changed in a scene, for example new buildings, tents, vehicles or construction sites. We also validated our method on two standard matching problems - stereo matching and optical flow on two standard datasets KITTI and Sintel~\cite{Butler2012}, where we obtained results comparable to the state-of-the-art. The resulting classifier typically obtains smooth results even
without any form of regularization, unlike other approaches, where this step is very crucial to remove noise and handle texture-less regions. A similar idea has been independently developed at the same time in~\cite{Zbontar14}, where the stereo matching cost is learnt using Convolutional Neural Networks.

\section{Designing the matching classifier}

First we show, that the problems of stereo matching, optical flow or pixel-wise change detection, are all conceptually the same. For stereo matching we would like to design a classifier, that predicts for each pixel in an input image the likelihood, that it matches the pixel in the reference shifted by a disparity. A properly designed classifier needs to be translation covariant; if we shift a reference image to any side by any number of pixels, the estimated disparities should change by the same number of pixels. Intuitively, this property can be satisfied by just the binary classifier, which would predict, whether two candidate pixels match or not. Even though the original stereo problem is multi-label, the symmetries of the classifier reduce the learning to a much simpler 2-label problem. The same conclusion can be drawn for the optical flow problem; the classifier should be translation covariant in both x and y dimensions, and thus should be modelled as a binary classifier. For pixel-wise change detection the images are already aligned and the algorithm should only determine, whether each pair of corresponding pixels match or not, which leads to the same binary problem with much larger range of changes in the visual appearance.

Let $I_1$ be the reference image and $I_2$ the image we match. As we already concluded, our goal is to learn a classifier $H(I_1, I_2, x_1, x_2)$, that predicts whether pixels $x_2 \in I_2$ matches to the pixel $x_1 \in I_1$ in the reference image. In a standard feature matching approach it is typically insufficient to match only the individual pixel colours, and thus features evaluated over a local neighbourhood (patch) are typically matched instead. The contextual range (size of patch) required to successfully match pixels is typically dataset-dependent; too small patches might contain insufficient information and in a too large patch all important information might get lost. For our classifier, we would like to learn the size of the context automatically. If the range of the context was determined by the rectangle $r$ centered at the pixels $x_1 \in I_1$ and $x_2 \in I_2$, in the ideal conditions the feature representations evaluated over each rectangle $\Phi^r(I_1, x_1)$ and $\Phi^r(I_2, x_2)$ would be nearly identical, i.e. the difference of the feature representation is close to a $0$-vector. In the presence of an unknown consistent systematic error, such as the second camera is consistently darker, or the first picture is always taken in summer and second scene in winter, this difference follows certain patterns, which should be learnable from the data. Thus, we want to model the matching classifier as the function of the difference of feature representations $\Phi^r(I_1, x_1) - \Phi^r(I_2, x_2)$. If the likelihood of a systematic error is the same for both images, we can alternatively use the absolute difference. The confidence of the classifier should significantly increase, if we use contextual information over multiple scales, the feature representations of multiple, not necessarily centered, rectangles surrounding the pixel. During the learning process, our classifier should determine the importance of different scales and contextual ranges automatically. Thus, the input vector will consist of the concatenation of differences of feature representations:
\begin{equation}
\Phi(I_1, I_2, x_1, x_2) = concat_{r \in R}(\Phi^r(I_1, x_i) - \Phi^r(I_2, x_2)),
\end{equation}
where $concat(.)$ is the concatenation operator and $R$ is the fixed set of randomly sampled rectangles surrounding the pixels being matched.

\subsection{Related approaches}
The contextual matching feature representation for learning the matching function is largely motivated by a successful application of similar representations~\cite{ShottonWRC06,shotton08,hierarchicalcrf,Shotton11} to the semantic segmentation problem. For this particular task, the direct concatenations of the feature vectors of the surrounding pixels~\cite{shotton08,Shotton11} or the concatenation of the feature representations of the rectangles~\cite{ShottonWRC06,hierarchicalcrf} surrounding the pixel have been previously successfully applied. The restricted forms of representations are used mainly, because they can be efficiently evaluated either by a direct look-up or using the integral images. The main difference between semantic classifiers and our matching classifier is, that the semantic classifier finds the mapping between the image patch and the semantic label; on the other hand for matching problem we learn the space of possible visual appearance changes the matching function should be invariant to.

\subsection{Feature representations for stereo matching}
The most common rectangle representation for a semantic classifier is the bag-of-words representation~\cite{ShottonWRC06,hierarchicalcrf}. Each individual dimension for such representation can be calculated using the integral images, built for each individual visual word. This leads to a very large memory consumption during training, and thus requires a sub-sampling (typically $4\times4$ - $10\times10$, depending on the amount of data) of the integral images. The sub-sampling by a factor $k$ restricts the location and the size of the rectangles to be the multipliers of $k$. This is not a big problem for the semantic segmentation, where it leads to noisy objects boundaries, because it could be handled using an appropriate regularization~\cite{hierarchicalcrf}. However, for stereo matching it would limit our matching precision. Thus, we combine the sub-sampled bag-of-words representation of each rectangle with another representation, which contains the average feature response of a low-dimensional non-clustered feature. Each dimension for any rectangle can also be calculated using integral images, but requires only one integral image per dimension, and thus does not require sub-sampling. Combining robust powerful imprecise bag-of-words representation with a rather weak but precise average-feature representation allows us to train both a strong and smooth classifier.

\subsection{Training the classifier}

We use a standard AdaBoost framework~\cite{friedman00additive} to learn the binary classifier $H(\Phi(I_1, x_1) - \Phi(I_2, x_2))$. Positive training samples correspond to each matching pairs for pixels. Negatives are randomly sampled from the overwhelmingly larger set of non-matching pairs. Positives, respectively negatives, were weighted based on inverse ratio of occurrence of the positive and negative class. Thus, training became independent on the number of negatives used, once a sufficient number is reached. In practice we used $50\times$ more negatives. As weak classifiers we used decision stumps, defined as:
\begin{equation}
h(I_1, I_2, x_1, x_2) = a \delta((\Phi^r(I_1, x_1) - \Phi^r(I_2, x_2))_i > \theta) + b,
\end{equation}
where $\Phi^r(.)_i$ is the $i-th$ dimension of a feature representation over rectangle $r$ and $\theta$, $a$ and $b$ are the parameters of the weak classifier. The final classifier is defined as a sum of weak classifiers:
\begin{equation}
H(I_1, I_2, x_1, x_2) = \sum_{m=1}^M h^m(I_1, I_2, x_1, x_2),
\end{equation}
where $h^m$ is the $m$-th weak classifier. The most discriminant weak classifier minimizing exponential loss is iteratively found in each round of boosting by randomly sampling dimensions of the feature representations $\Phi^r(I_1, x_i) - \Phi^r(I_2, x_2)$, a brute force search for the optimal $\theta$ and close form~\cite{friedman00additive} evaluation of the $a$ and $b$ parameters of the weak classifier. We refer to~\cite{friedman00additive} for more details. The large dimensional feature representations are not kept in memory, but rather evaluated as needed by the classifier using the integral images.

The learning procedure automatically learns, which features are rare and significant, such that they must be matched precisely; and which features are common, for which the regions they belong to must be matched instead. Furthermore, the learning algorithm determines the importance of different scales and the space of appearance changes, it should be invariant to. Unlike in coarse-to-fine approaches, the matching function is found jointly over all scales together.

The prediction is done by evaluating the classifier for each pair of matching candidates independently. For stereo estimation the response for pixel $x$ and disparity $d$ is $H(x, d) = H(I_1, I_2, x, x - (d, 0))$, for optical flow the response for flow $f = (f_x, f_y)$ is $H(x, f) = H(I_1, I_2, x, x + f)$ and for pixel-wise change detection it is $H(x) = H(I_1, I_2, x, x)$.

\subsection{Regularization}
For stereo and optical flow problems we adapt the fully connected CRF approach~\cite{Krahenbuhl11} minimizing energy:
\begin{equation}
E({\bf d}) = -\sum_{i \in I_2} H(x_i, d_i) + \sum_{i,j \in I_2, i \neq j} \psi_{ij}(d_i, d_j),
\end{equation}
where $-H(x_i, d_i)$ is the unary potential for pixel $x_i$ and the pairwise potentials $\psi_{ij}(d_i, d_j)$ take the form:
\begin{equation}
\psi_{ij}(d_i, d_j) = \mu_{ij}(d_i, d_j) \exp(-\frac{|C_i - C_j|^2}{2\sigma^2_{app}} - \frac{|x_i - x_j|^2}{2\sigma^2_{loc}}),
\end{equation}
where $C_i$ and $C_j$ are the colour (appearance) features for pixels $i$ and $j$, $x_i$ and $x_j$ are the locations and $\sigma^2_{app}$ and $\sigma^2_{loc}$ are the widths of the location and appearance kernels. The label compatibility function $\mu_{ij}(d_i, d_j)$ enforces local planarity and takes the form:
\begin{equation}
\mu_{ij}(d_i, d_j) = \exp(-\frac{|d_i - d_j - p_i\cdot(x_j - x_i)|^2}{2\sigma^2_{pln}}),
\end{equation}
where $\sigma^2_{pln}$ is the width of the plane kernel and $p_i = (p_i^1, p_i^2)$ are the coefficients of a fitted plane, obtained using RANSAC by maximizing:
\begin{eqnarray}
p_i = \arg \max_{p'} \sum_{j \in I_2} \bigg{(} \exp(-\frac{|C_i - C_j|^2}{2\sigma^2_{app}} - \frac{|x_i - x_j|^2}{2\sigma^2_{loc}})\nonumber\\
 \delta(|d_i - d_j - p'\cdot(x_j - x_i)| \leq \sigma)\bigg{)}~\label{ransac},
\end{eqnarray}
where $\sigma = 1px$ is the threshold and $\delta(.)$ Kronecker delta function. In each iteration the local plane is fitted for each pixel $x_i$ based on the current marginal beliefs by solving the optimization problem (\ref{ransac}) using RANSAC. In the next step, the marginals are updated for each pixel as described in~\cite{Krahenbuhl11}. We repeat this procedure till converges or till our maximum number of allowed iterations is reached.

\subsection{Implementation details}

In our implementation of the classifier we used the $4$ bag-of-words representations using texton~\cite{Malik01}, SIFT~\cite{lowe04}, local quantized ternary patters~\cite{Hussain2012} and self-similarity features~\cite{Shechtman07}. Each bag-of-words representation for each rectangle consisted of a $512$ dimensional vector, encoding the soft-weighted occurrence of each visual word in the rectangle. The visual words (cluster centres) were obtained using k-means clustering. The soft weights for $8$ nearest visual words were calculated using a distance-based exponential kernel~\cite{Gemert08a}. Integral images for the bag-of-words representations were sub-sampled by a factor of $4\times4$. Additionally we used the average-feature representation, consisting of $17$ dimensional non-clustered texton features, containing average responses of $17$ filter banks. The final representation was a concatenation of $4$ bag-of-words and $1$ average-feature representation, both over $200$ rectangles (different to each other due to a different sub-sampling of these representations). The final feature vector - an input of the classifier - was $(4 \times 512 + 17) \times 200$-dimensional. The boosted classifier consisted of $5000$ weak classifier - decision stumps, as in~\cite{ShottonWRC06}. The difference of feature representations can be discriminative only if the rectangles in the left and right images are of the same size. This could be a problem if the shifted rectangle is differently cropped by the edge of an image. To avoid this, we simply detect such problem and update the cropping of another rectangle accordingly. Prior to regularization, a validation step of inverse matching from the reference frame to the frame matched was performed, to handle occluded regions. Classifier responses for the pixels that, did not pass the validation step, have been ignored in the regularization.

\section{Experiments}
\label{sec_experiments}
We evaluated the classifier for three matching problems - outdoor KITTI~\cite{Geiger2012CVPR} stereo dataset, Sintel optical flow dataset and our new TimeLapse change detection dataset.

\subsection{KITTI data set}
\begin{figure*}
\centering
\begin{tabular}{cccc}
\includegraphics[width=0.235\linewidth, height=0.077\linewidth]{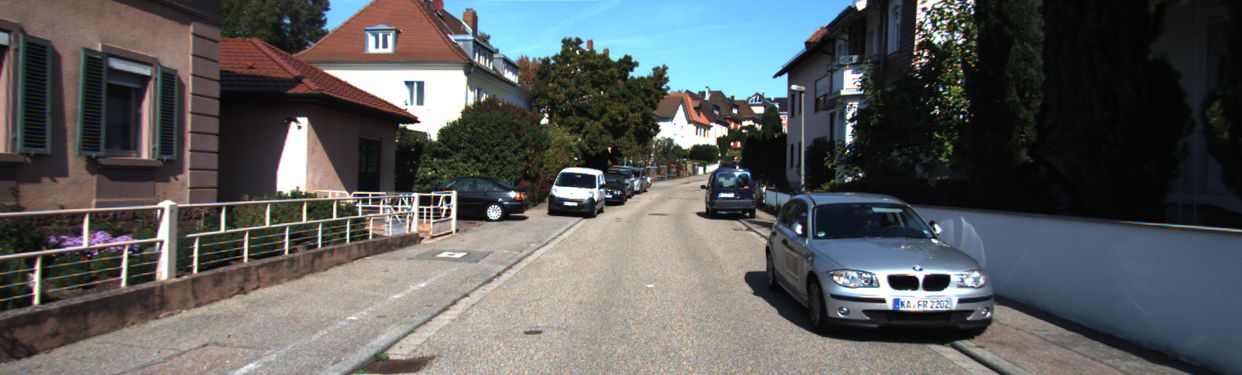} &
\includegraphics[width=0.235\linewidth, height=0.077\linewidth]{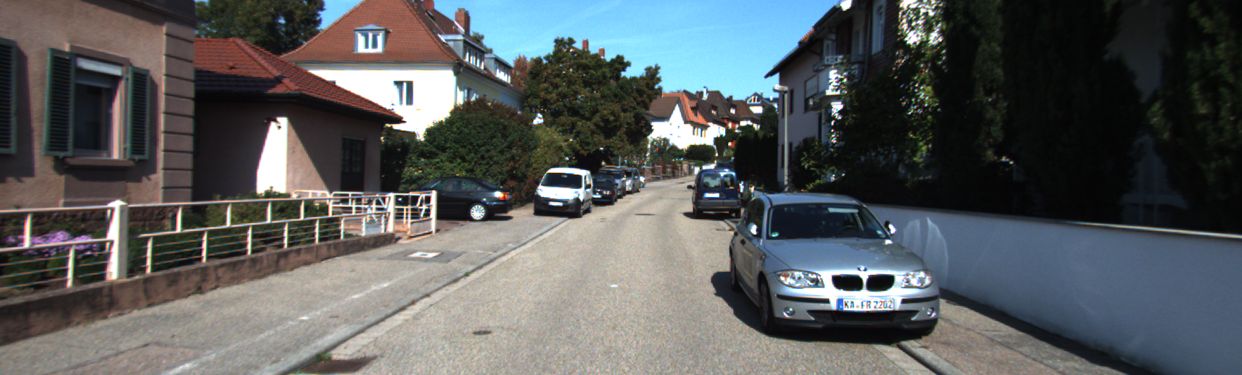} &
\includegraphics[width=0.235\linewidth, height=0.077\linewidth]{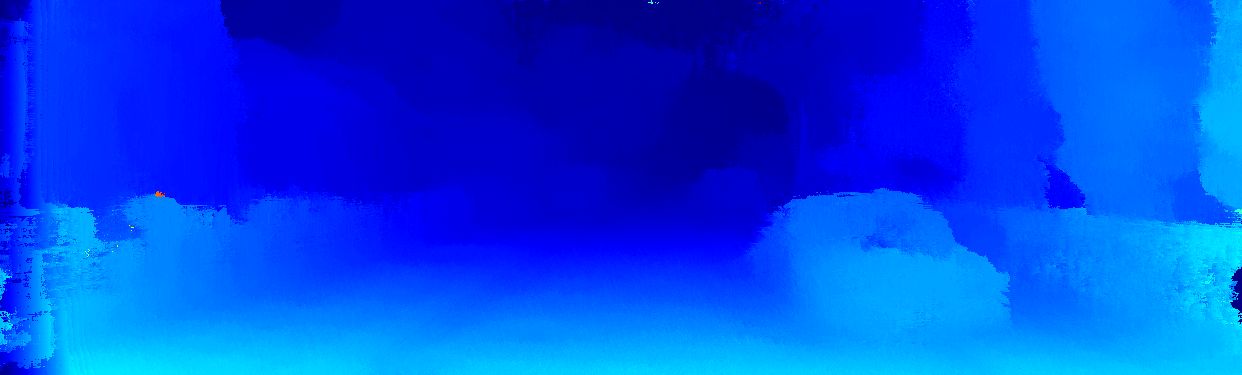} &
\includegraphics[width=0.235\linewidth, height=0.077\linewidth]{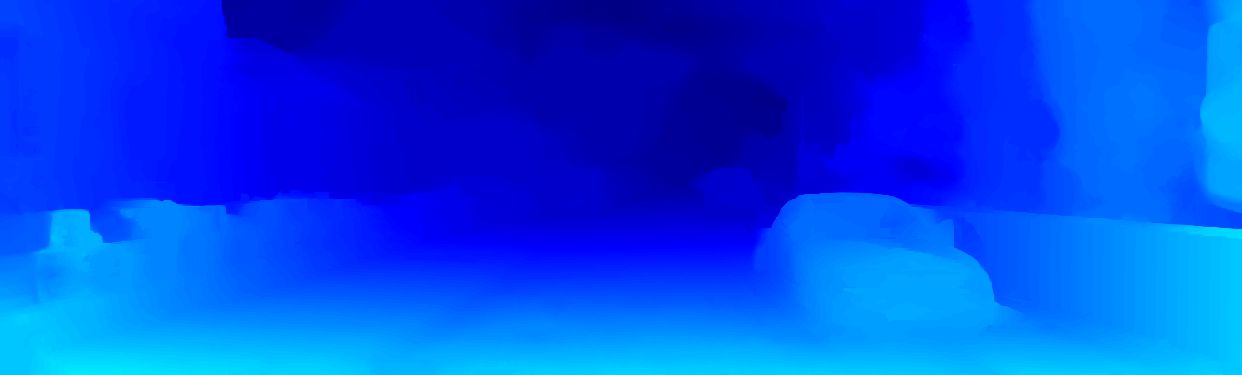} \\

\includegraphics[width=0.235\linewidth, height=0.077\linewidth]{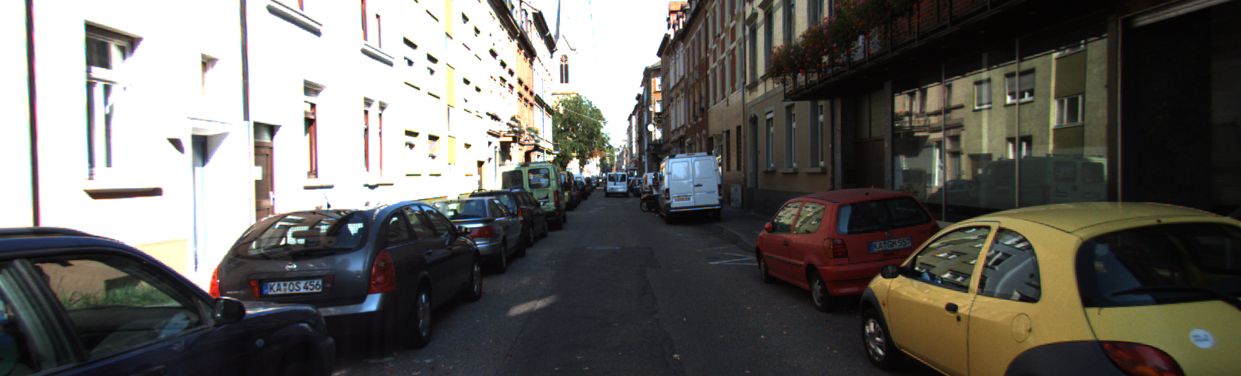} &
\includegraphics[width=0.235\linewidth, height=0.077\linewidth]{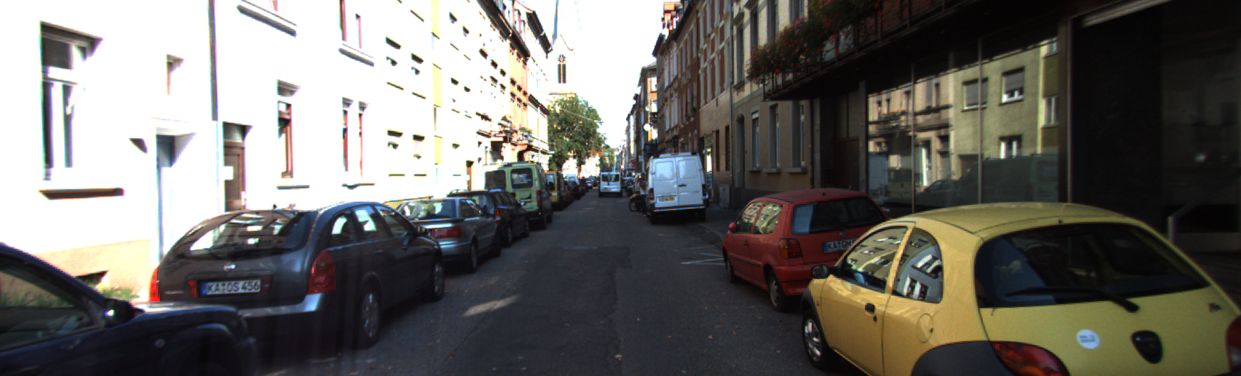} &
\includegraphics[width=0.235\linewidth, height=0.077\linewidth]{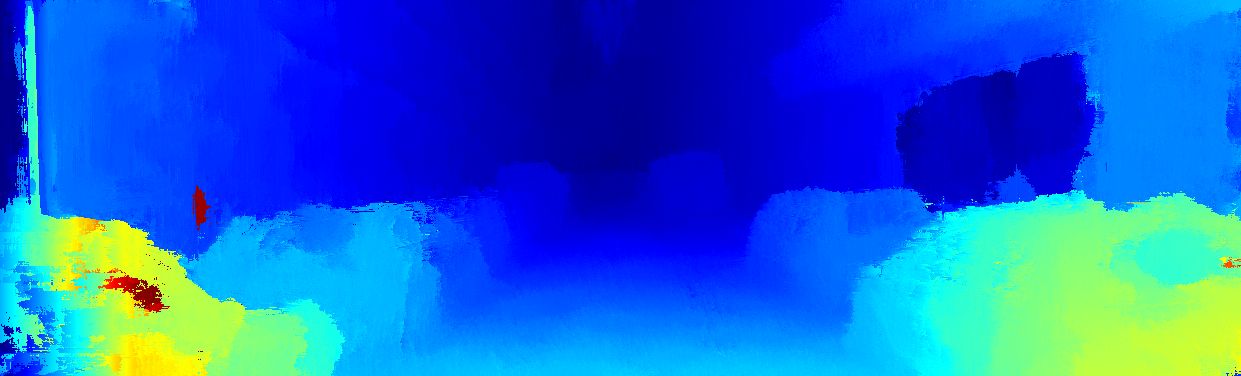} &
\includegraphics[width=0.235\linewidth, height=0.077\linewidth]{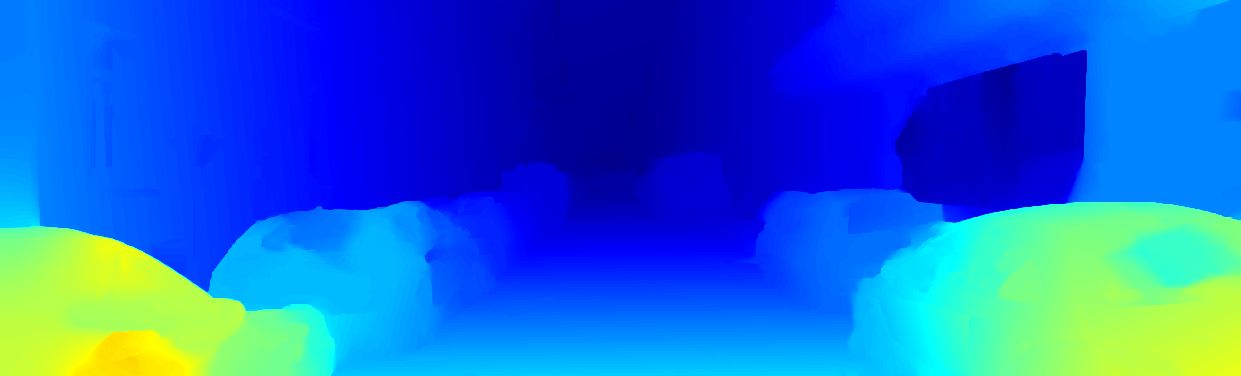} \\

\includegraphics[width=0.235\linewidth, height=0.077\linewidth]{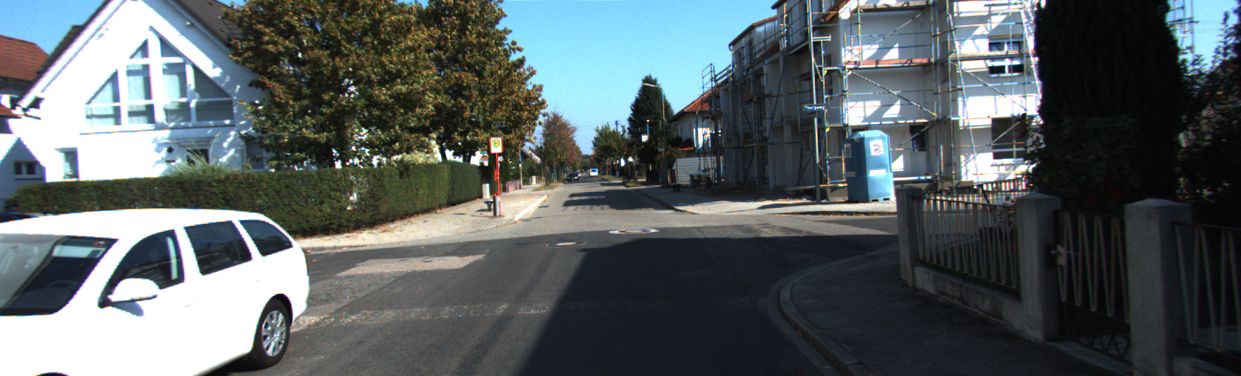} &
\includegraphics[width=0.235\linewidth, height=0.077\linewidth]{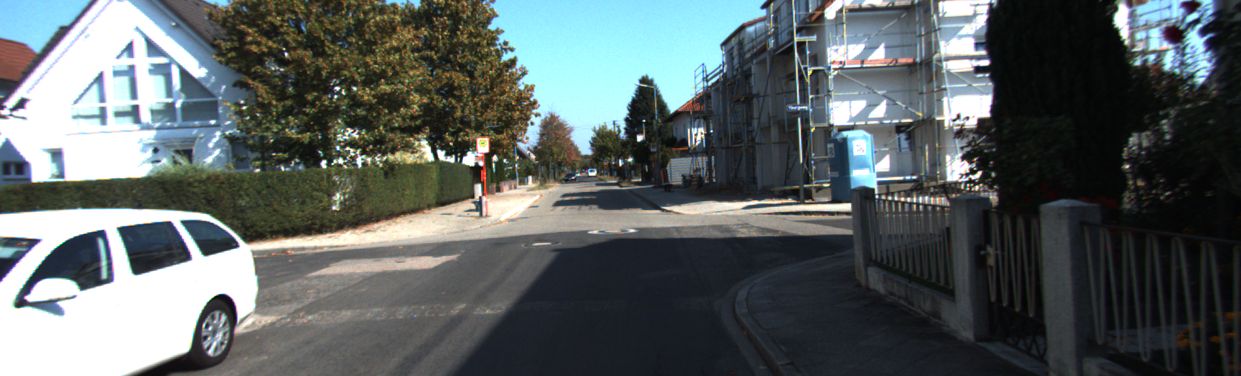} &
\includegraphics[width=0.235\linewidth, height=0.077\linewidth]{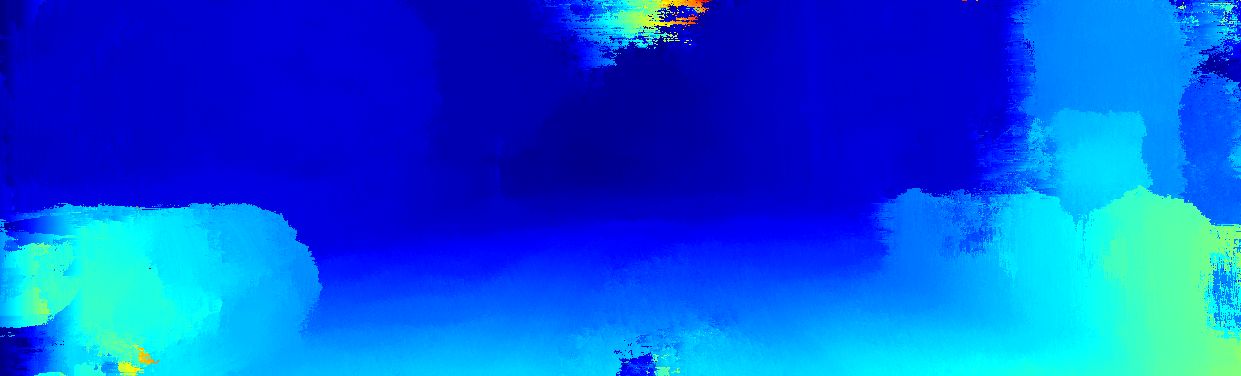} &
\includegraphics[width=0.235\linewidth, height=0.077\linewidth]{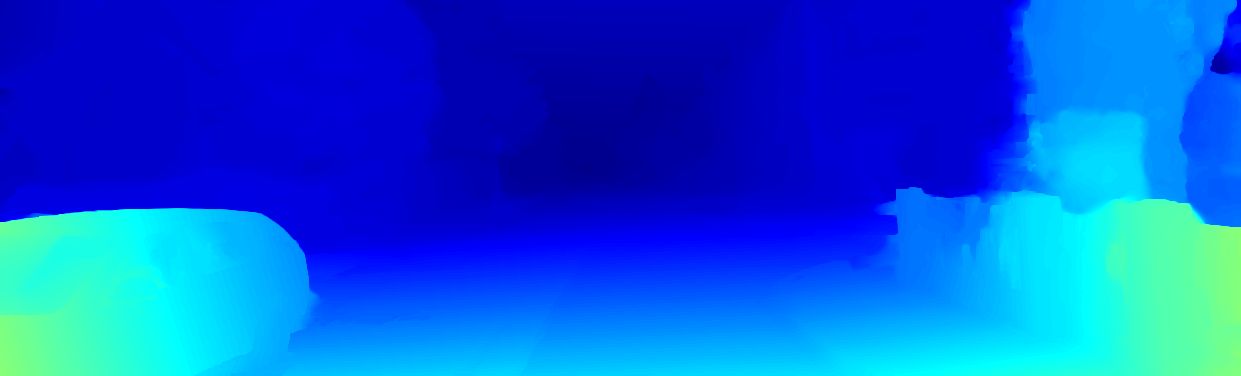} \\

\includegraphics[width=0.235\linewidth, height=0.077\linewidth]{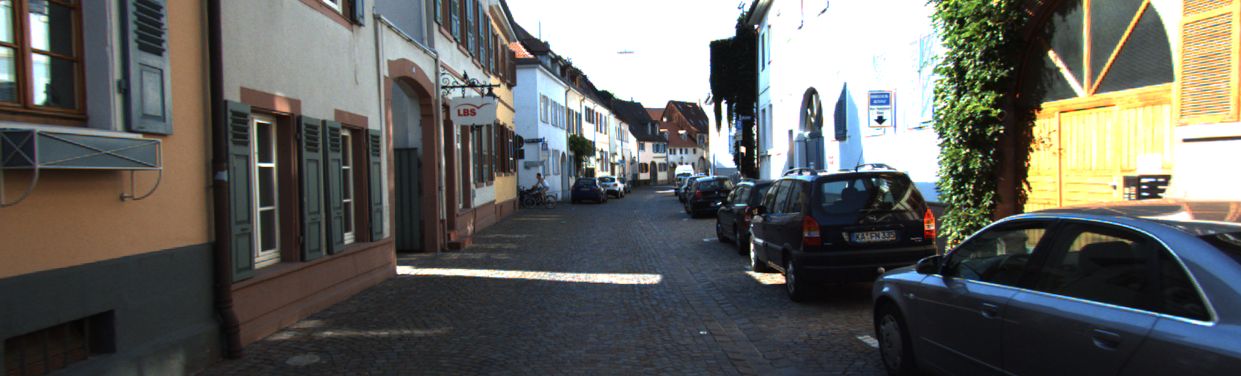} &
\includegraphics[width=0.235\linewidth, height=0.077\linewidth]{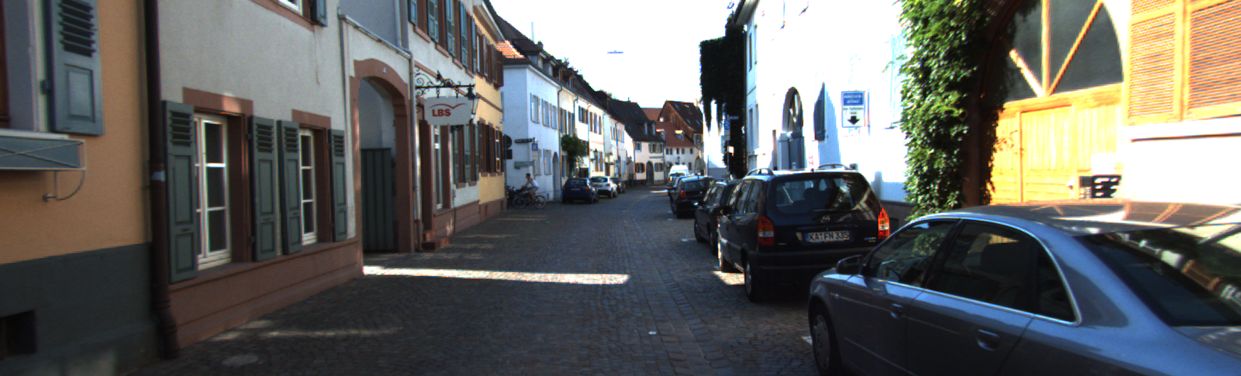} &
\includegraphics[width=0.235\linewidth, height=0.077\linewidth]{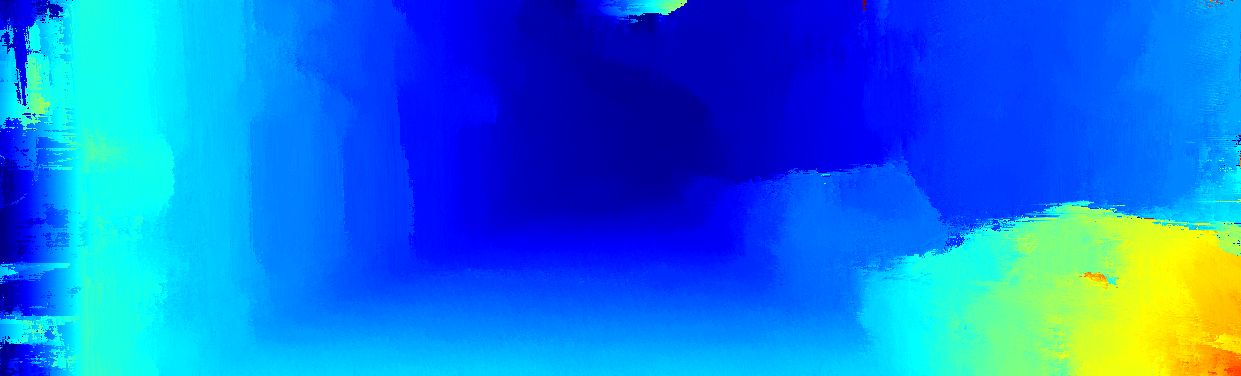} &
\includegraphics[width=0.235\linewidth, height=0.077\linewidth]{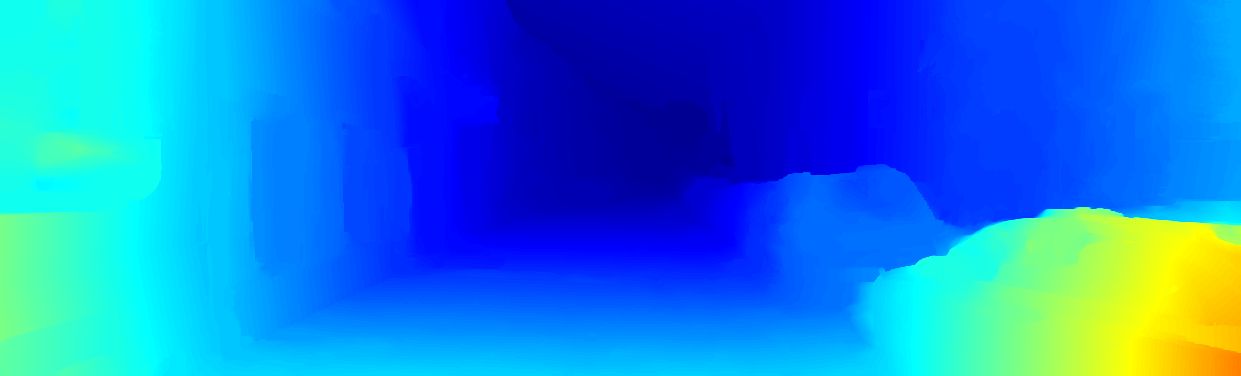} \\

\includegraphics[width=0.235\linewidth, height=0.077\linewidth]{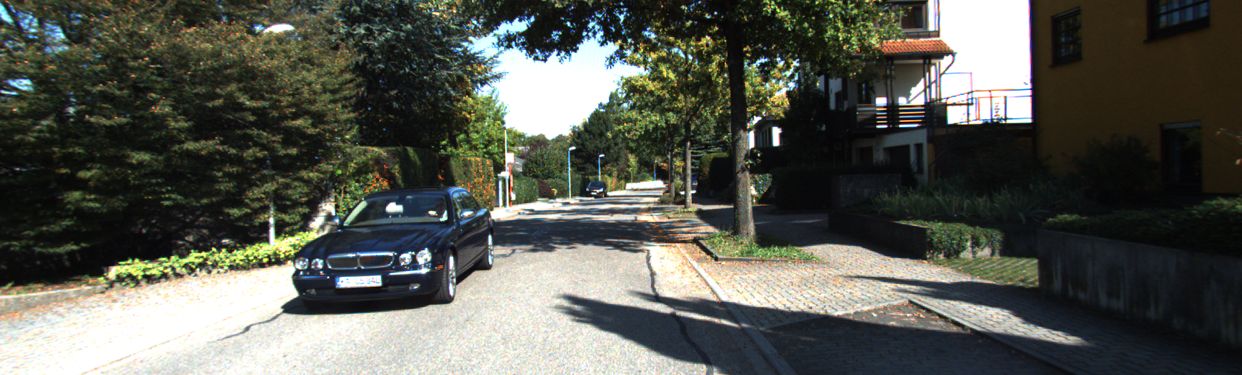} &
\includegraphics[width=0.235\linewidth, height=0.077\linewidth]{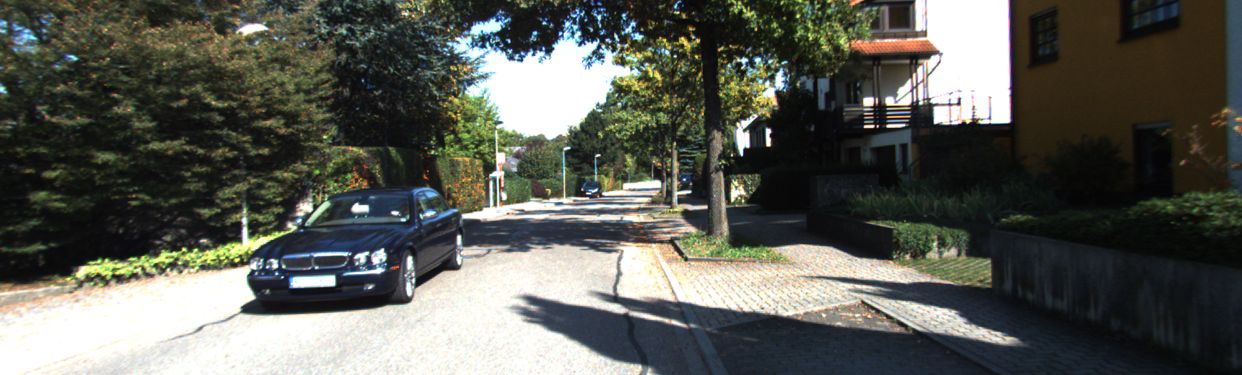} &
\includegraphics[width=0.235\linewidth, height=0.077\linewidth]{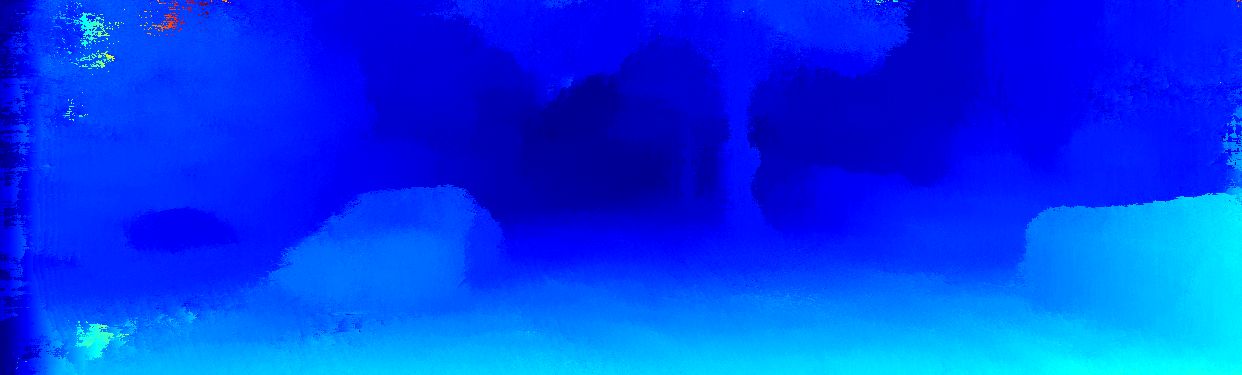} &
\includegraphics[width=0.235\linewidth, height=0.077\linewidth]{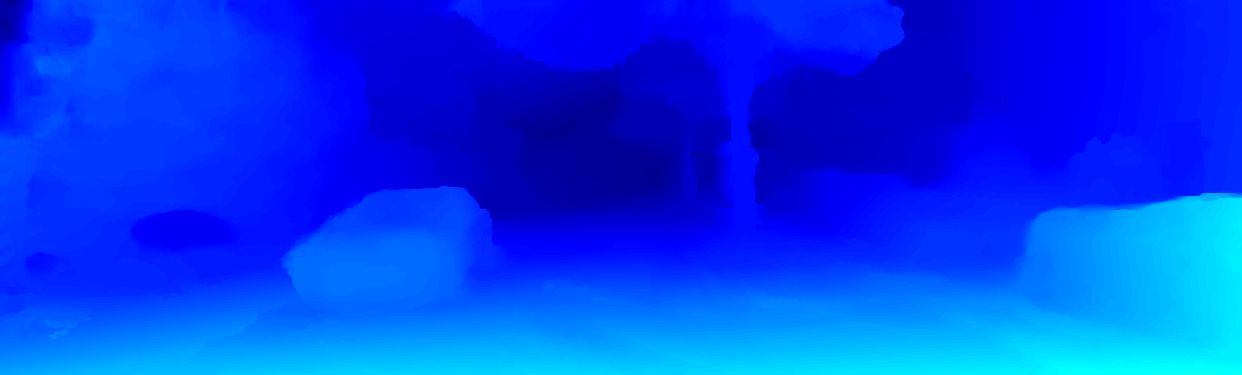} \\

\includegraphics[width=0.235\linewidth, height=0.077\linewidth]{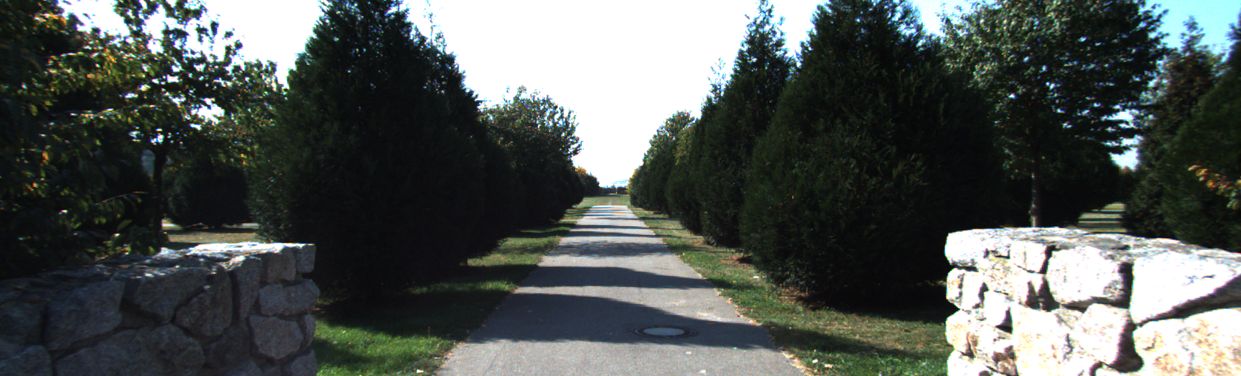} &
\includegraphics[width=0.235\linewidth, height=0.077\linewidth]{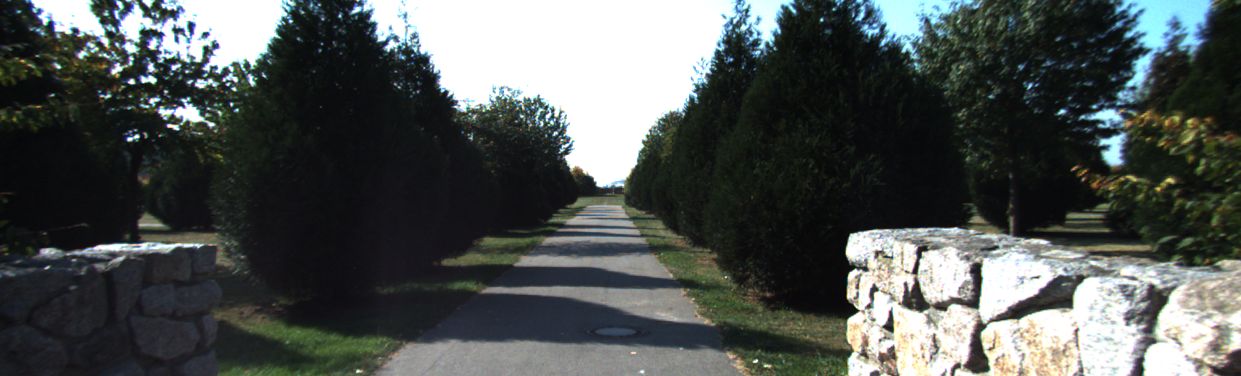} &
\includegraphics[width=0.235\linewidth, height=0.077\linewidth]{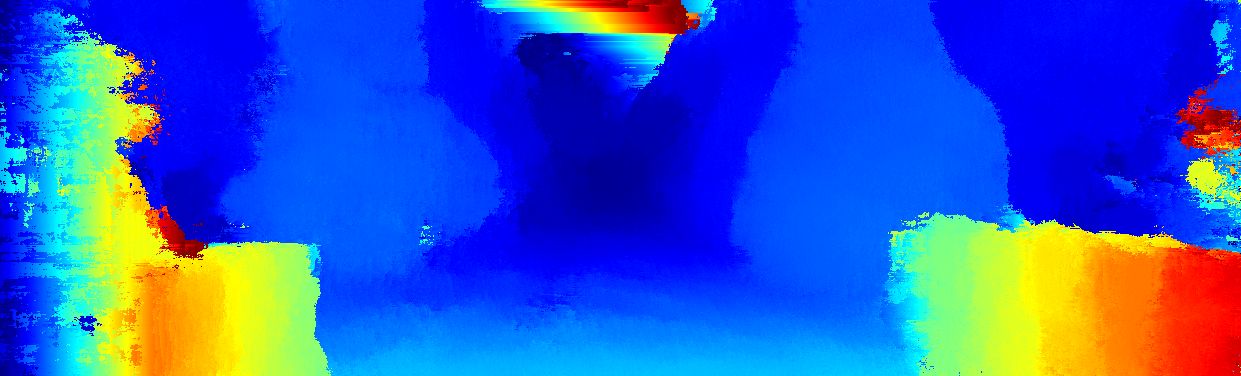} &
\includegraphics[width=0.235\linewidth, height=0.077\linewidth]{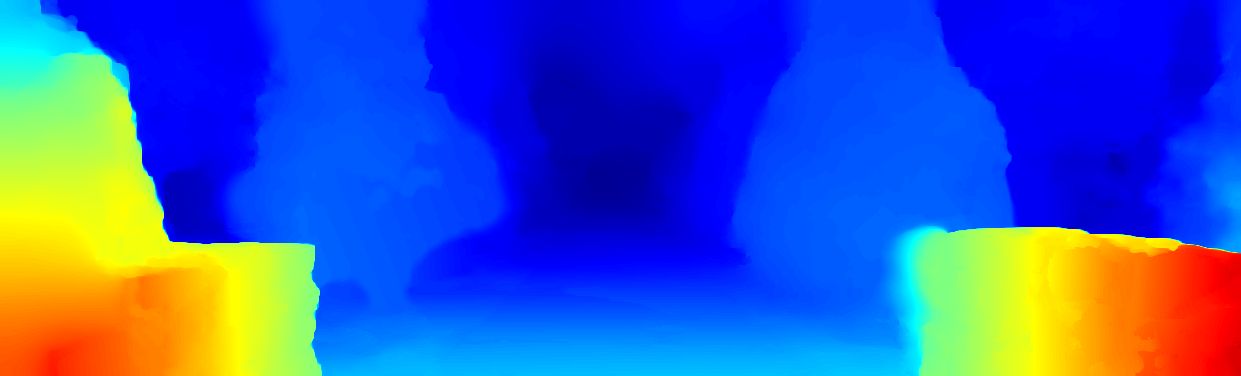} \\

\includegraphics[width=0.235\linewidth, height=0.077\linewidth]{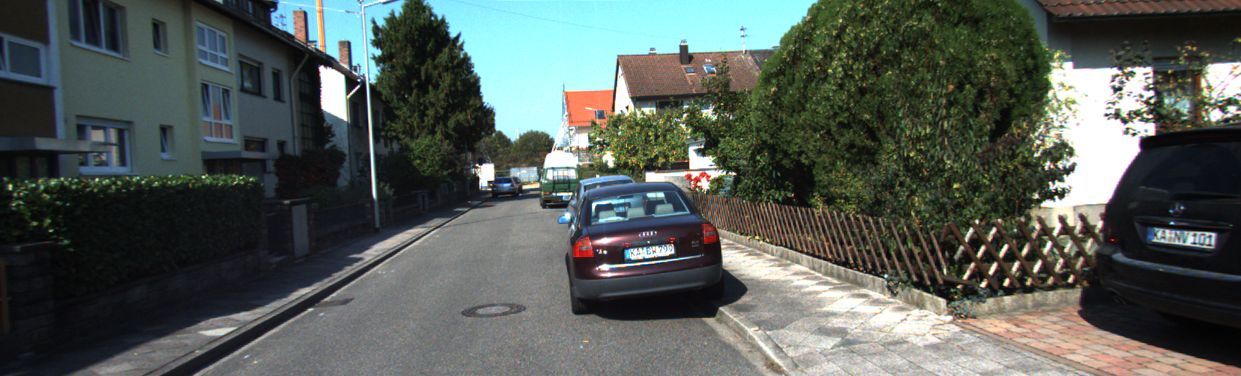} &
\includegraphics[width=0.235\linewidth, height=0.077\linewidth]{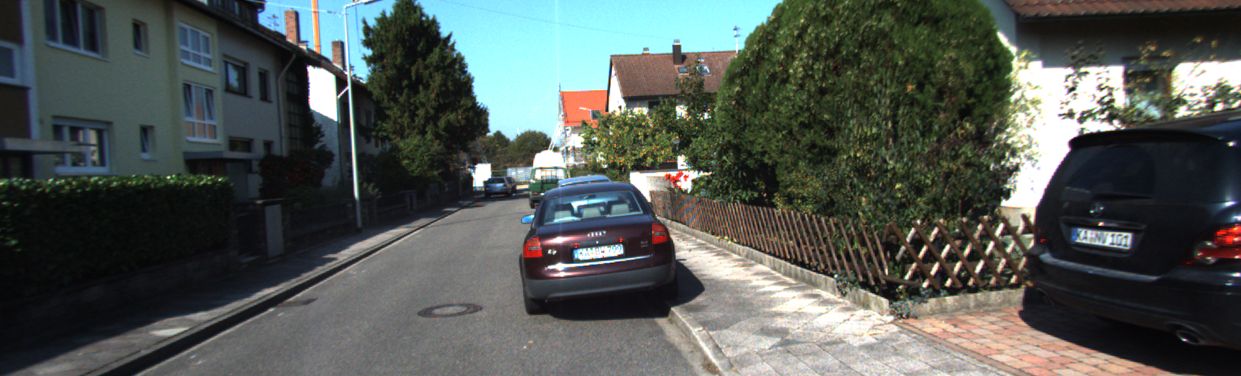} &
\includegraphics[width=0.235\linewidth, height=0.077\linewidth]{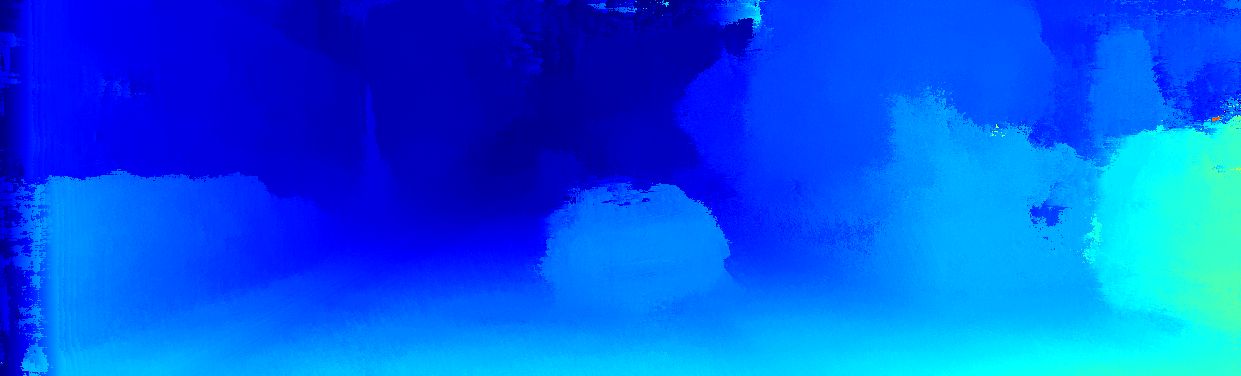} &
\includegraphics[width=0.235\linewidth, height=0.077\linewidth]{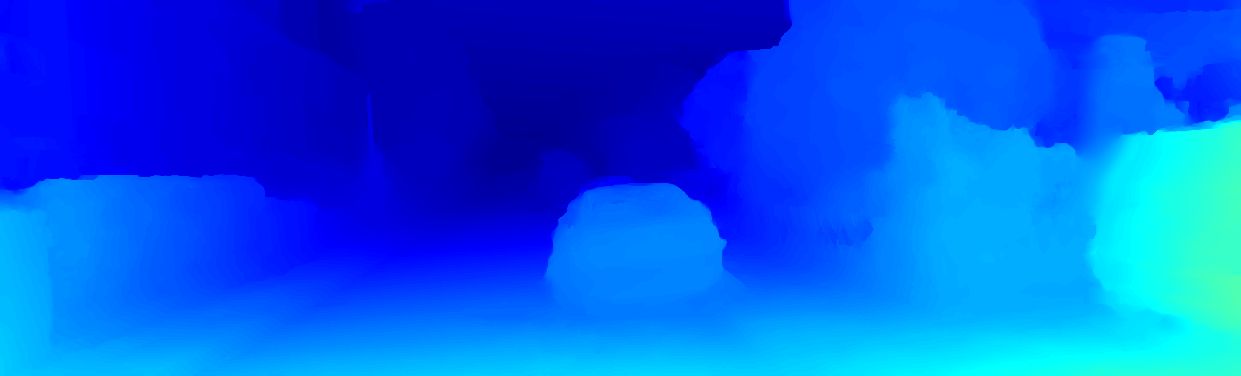} \\

\includegraphics[width=0.235\linewidth, height=0.077\linewidth]{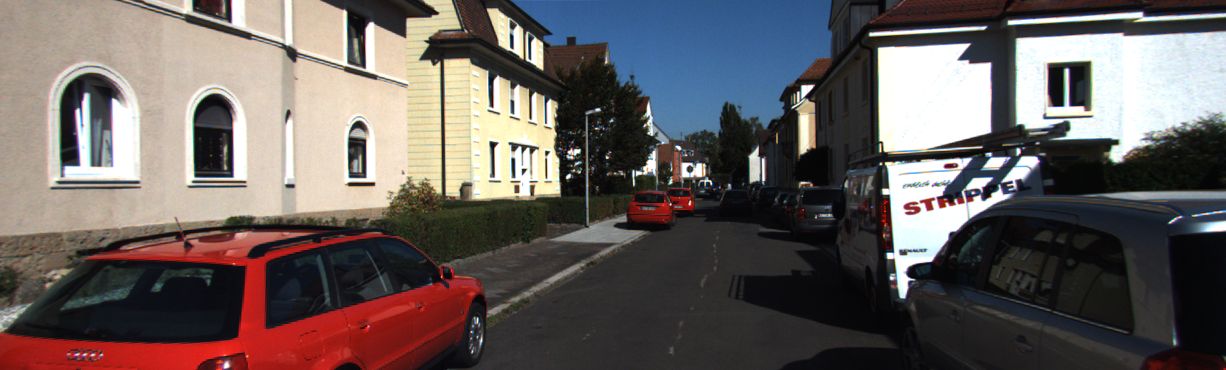} &
\includegraphics[width=0.235\linewidth, height=0.077\linewidth]{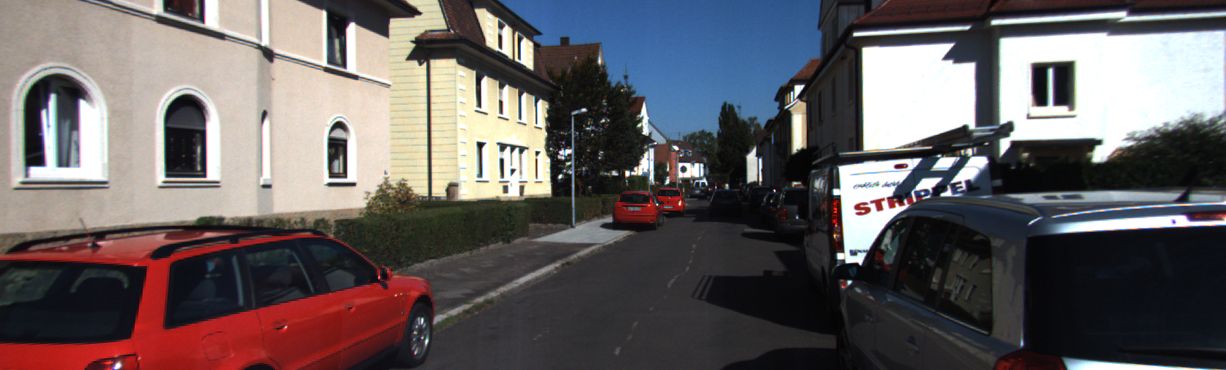} &
\includegraphics[width=0.235\linewidth, height=0.077\linewidth]{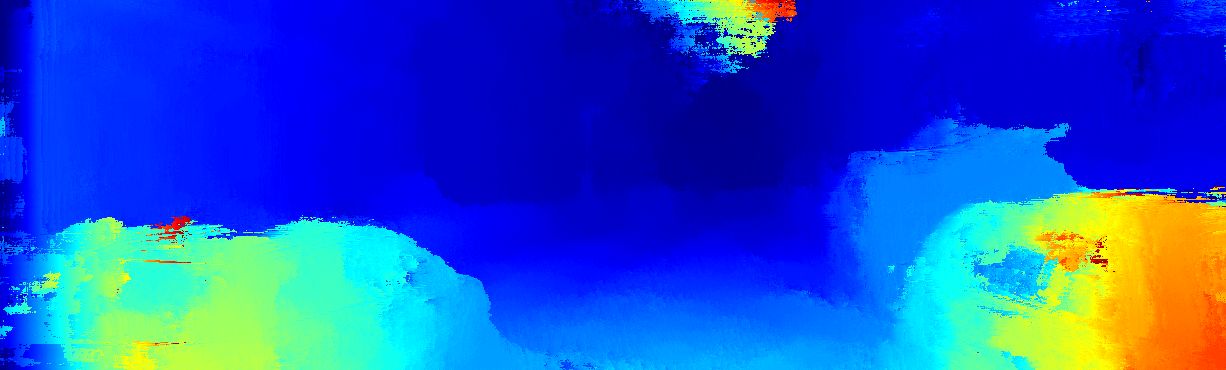} &
\includegraphics[width=0.235\linewidth, height=0.077\linewidth]{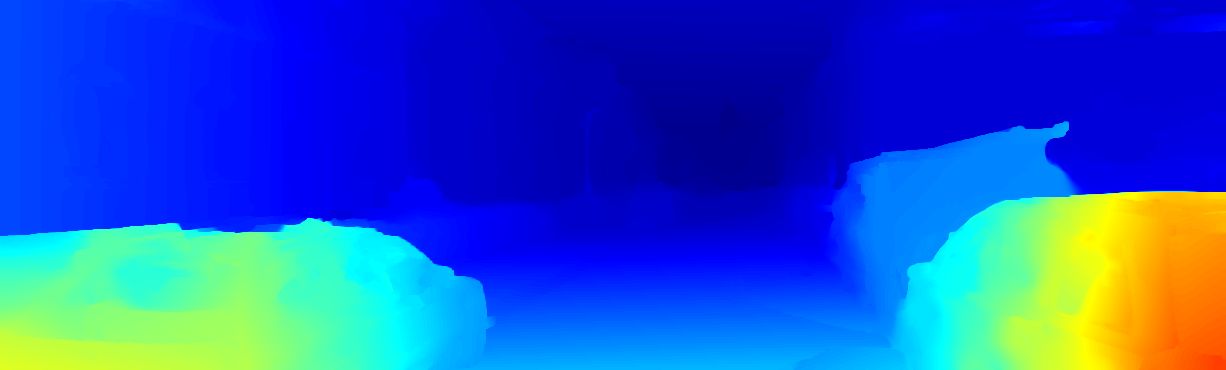} \\

\includegraphics[width=0.235\linewidth, height=0.077\linewidth]{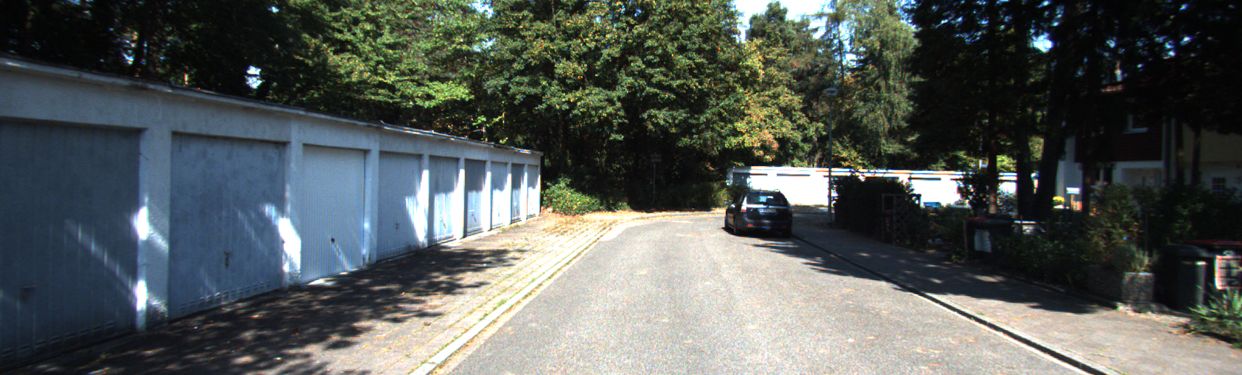} &
\includegraphics[width=0.235\linewidth, height=0.077\linewidth]{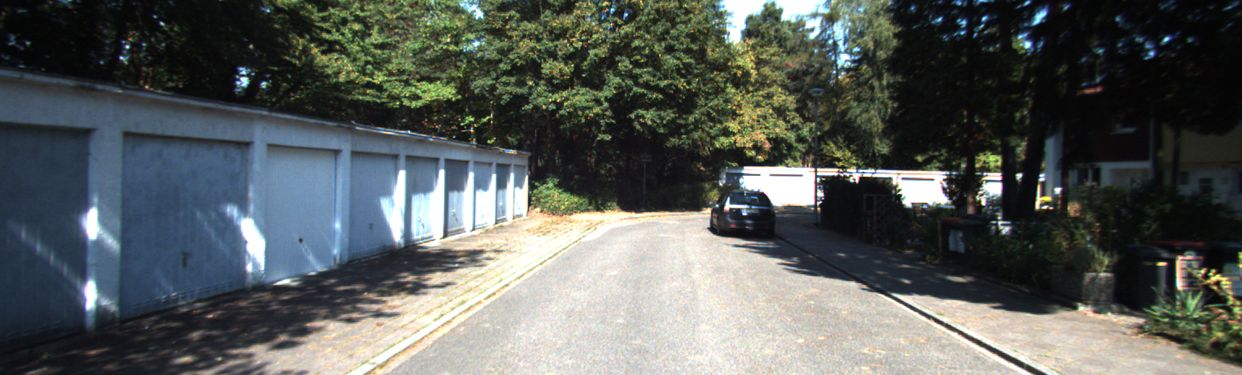} &
\includegraphics[width=0.235\linewidth, height=0.077\linewidth]{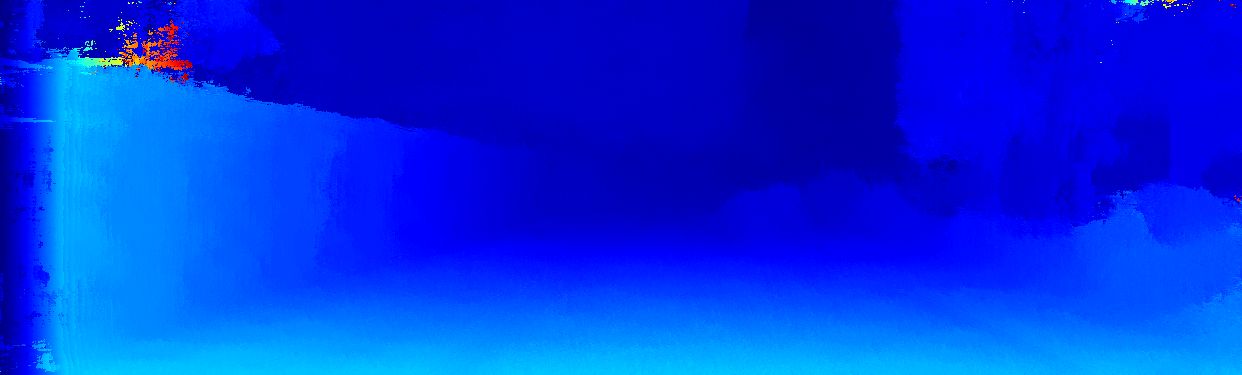} &
\includegraphics[width=0.235\linewidth, height=0.077\linewidth]{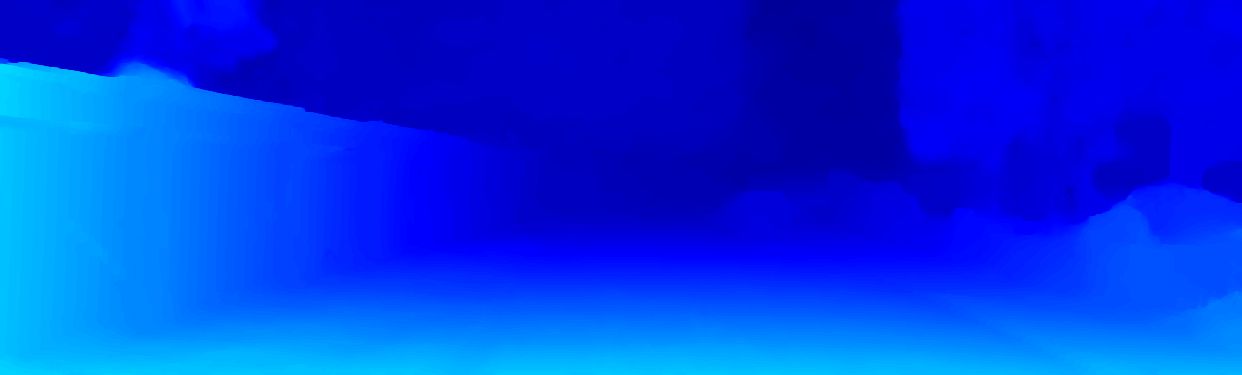} \\

\includegraphics[width=0.235\linewidth, height=0.077\linewidth]{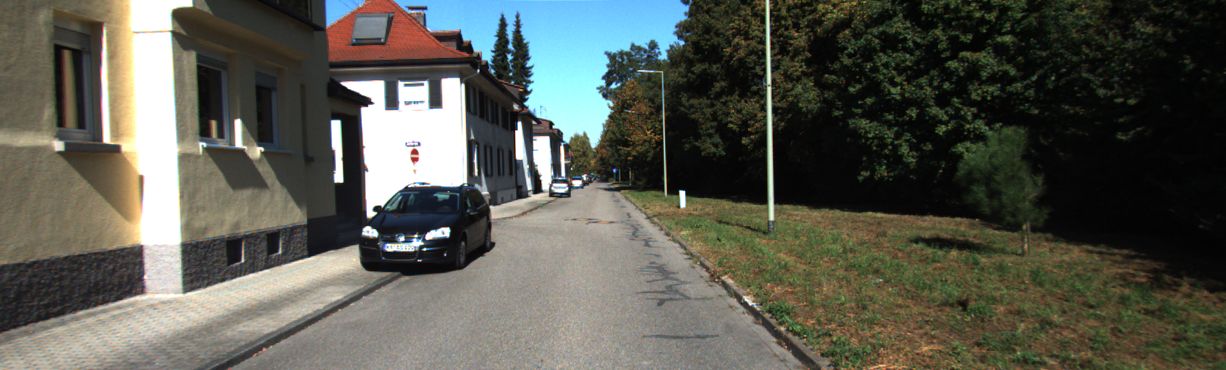} &
\includegraphics[width=0.235\linewidth, height=0.077\linewidth]{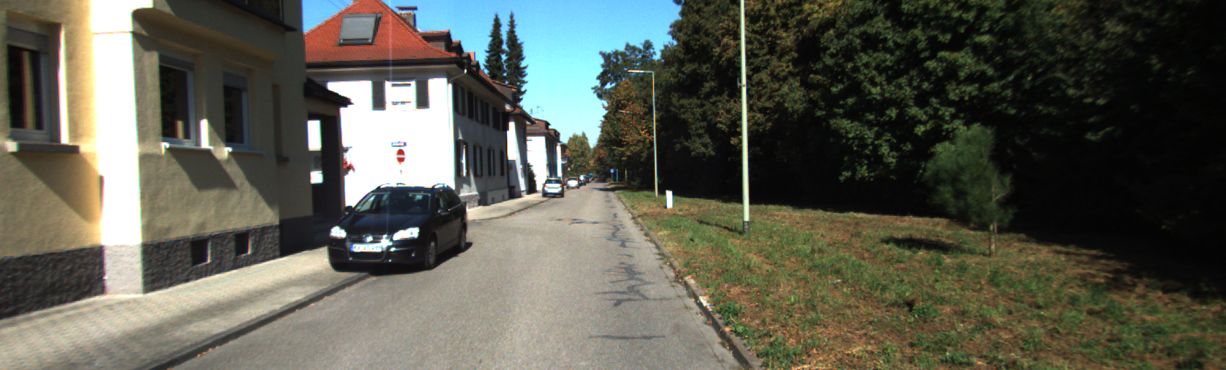} &
\includegraphics[width=0.235\linewidth, height=0.077\linewidth]{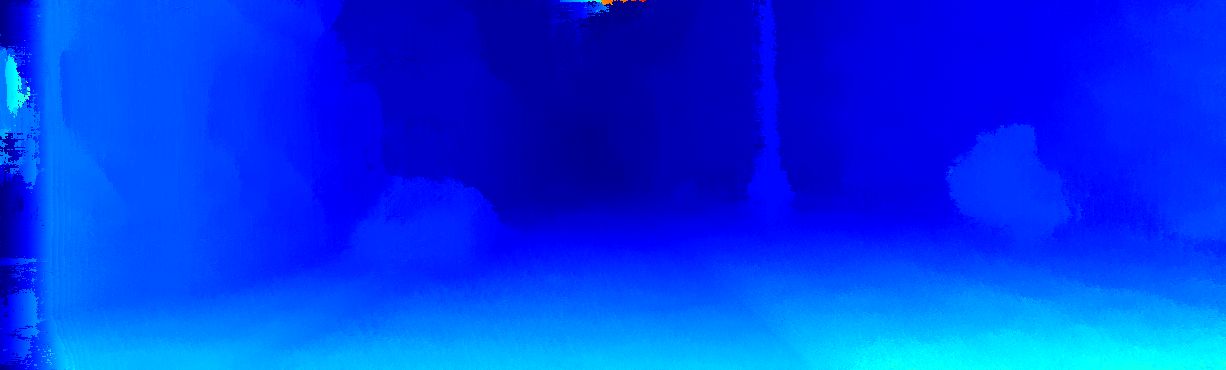} &
\includegraphics[width=0.235\linewidth, height=0.077\linewidth]{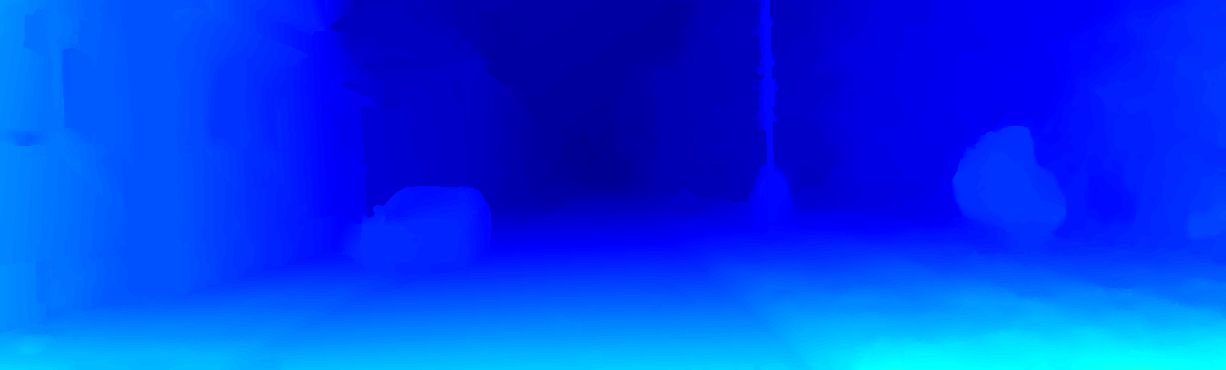} \\

\includegraphics[width=0.235\linewidth, height=0.077\linewidth]{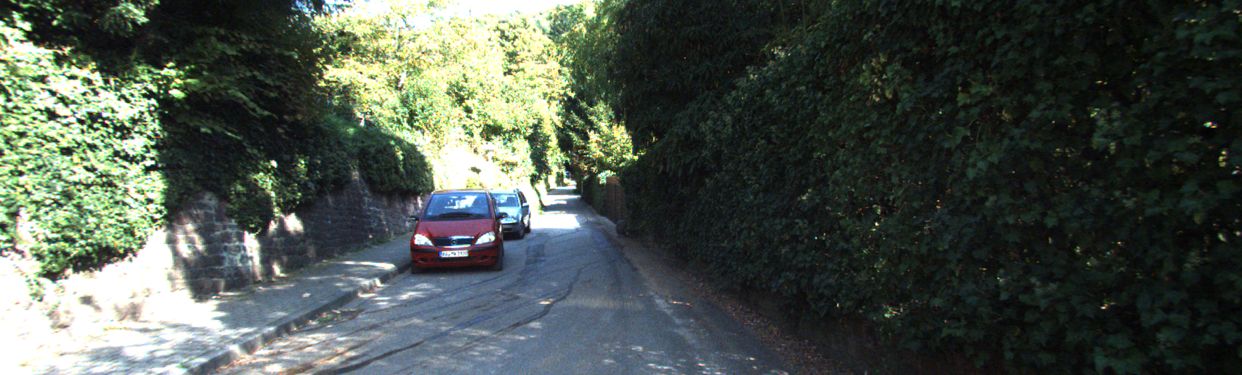} &
\includegraphics[width=0.235\linewidth, height=0.077\linewidth]{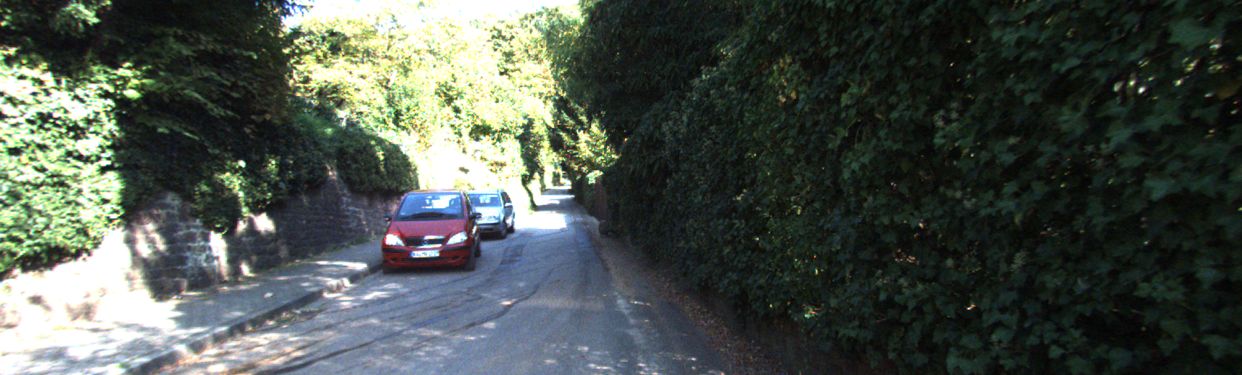} &
\includegraphics[width=0.235\linewidth, height=0.077\linewidth]{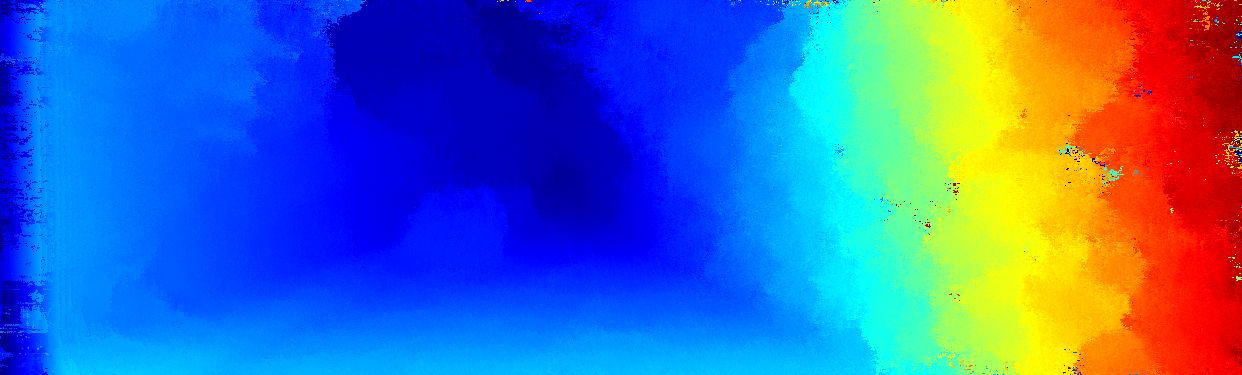} &
\includegraphics[width=0.235\linewidth, height=0.077\linewidth]{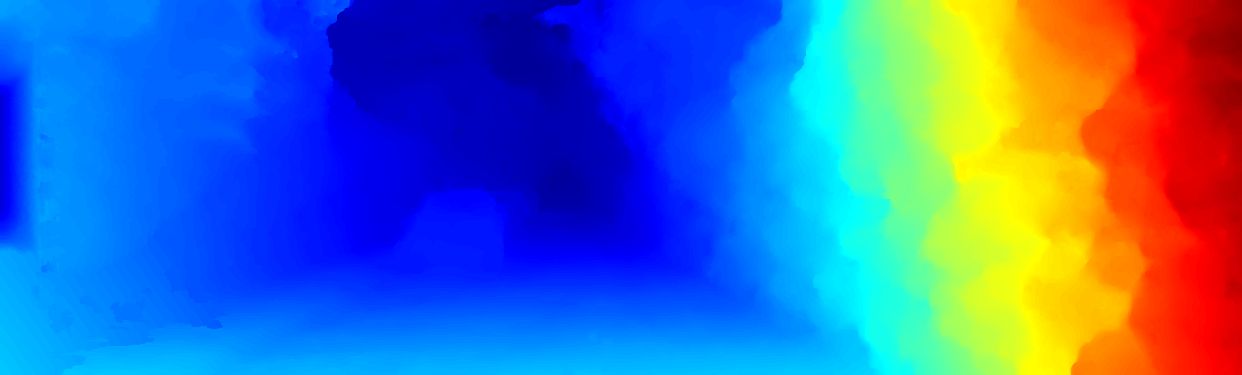} \\
Left Image & Right Image & Unary Result & Regularized Result
\end{tabular}
\includegraphics[width=1\linewidth]{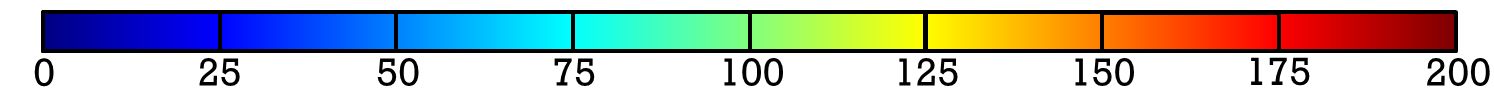}
 \small{\caption{Qualitative results on the KITTI dataset. Our classifier could successfully match pixels for a wide range of depths. Most of mismatches are due to the occlusion or for sky pixels, which were not used during training. Regularization further improved the results by inducing smoothness and local planarity. Reconstructed 3D point clouds for the same images could be found in the figure~\ref{KITTI_3D}.
\label{KITTI_res}}}
\end{figure*}

\begin{figure*}
\centering
\begin{tabular}{cccc}
\includegraphics[width=0.235\linewidth, height=0.077\linewidth]{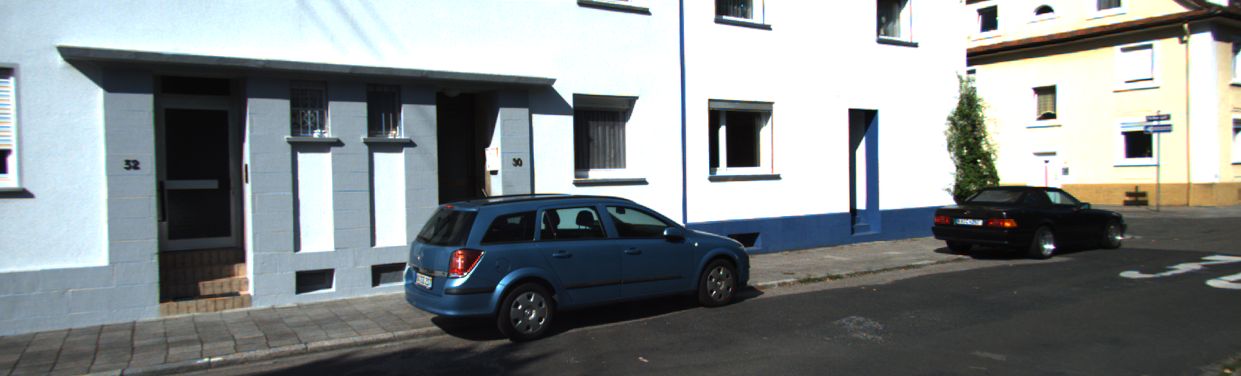} &
\includegraphics[width=0.235\linewidth, height=0.077\linewidth]{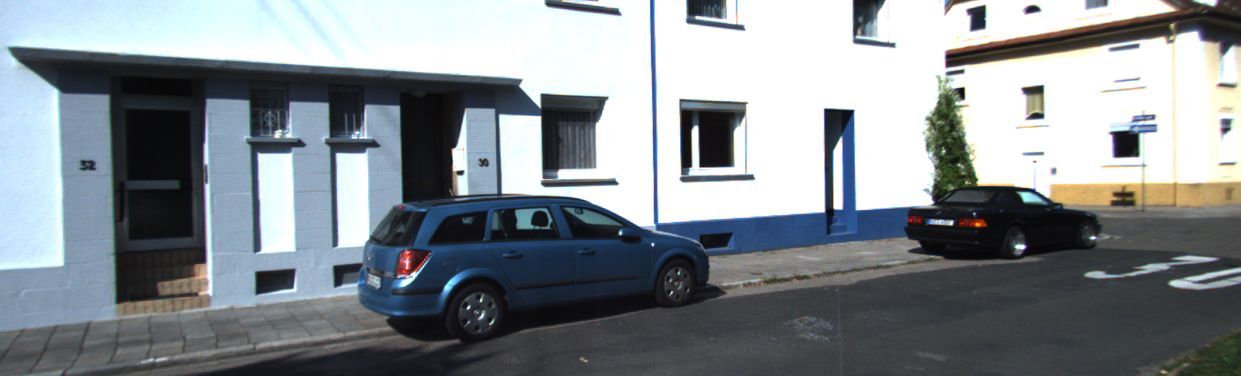} &
\includegraphics[width=0.235\linewidth, height=0.077\linewidth]{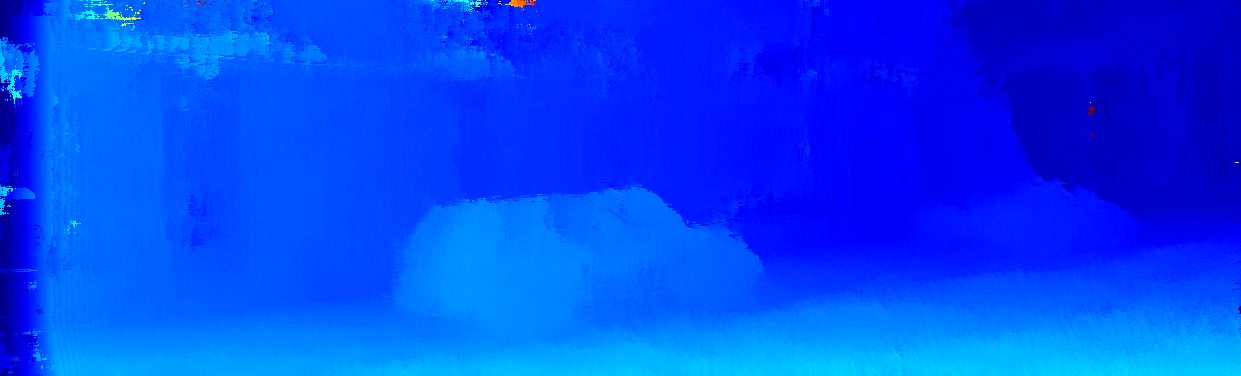} &
\includegraphics[width=0.235\linewidth, height=0.077\linewidth]{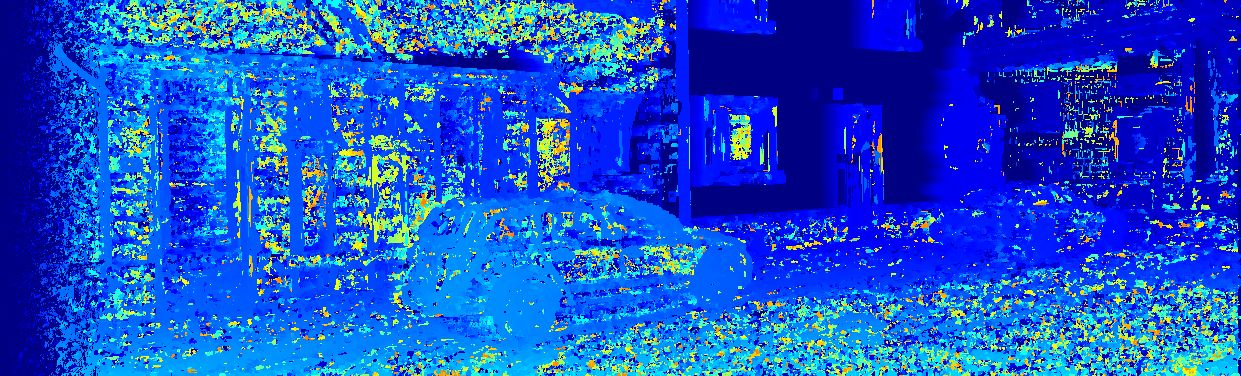} \\
Left image & Right image & Our Unary Result & Feature Matching Result
\end{tabular}
\small{\caption{Comparison of our unary result with standard feature matching classifier, obtained as an average of Sobel Filter matching~\cite{geiger2010efficient} and Census transform~\cite{zabih1994non}. Standard unary matching result is much more noisy and fails on texture-less surfaces, such as white walls.
\label{KITTI_comp}}}\end{figure*}

\begin{figure*}
\centering
\begin{tabular}{cccc}
\vspace{2mm}\includegraphics[width=0.22\linewidth, height=0.13\linewidth]{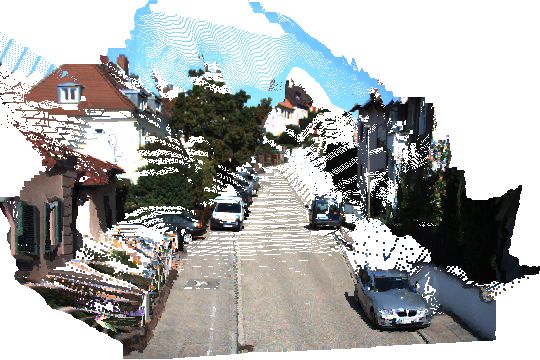} &
\includegraphics[width=0.22\linewidth, height=0.13\linewidth]{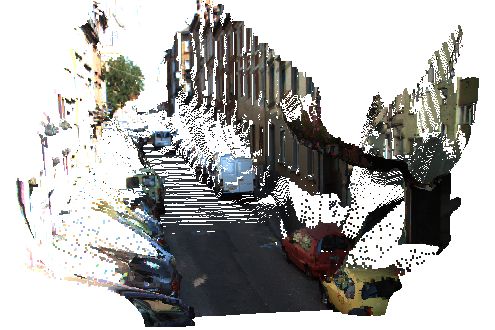} &
\includegraphics[width=0.22\linewidth, height=0.13\linewidth]{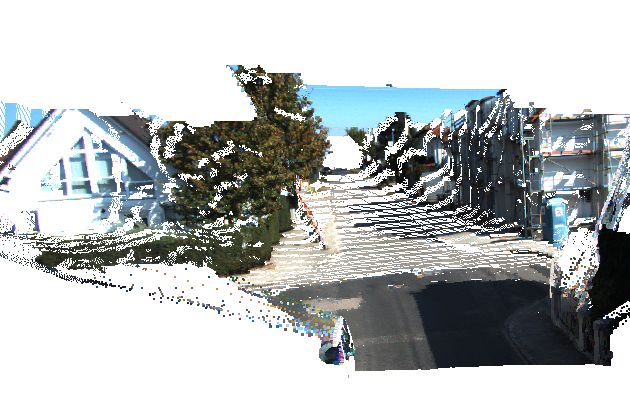} &
\includegraphics[width=0.22\linewidth, height=0.13\linewidth]{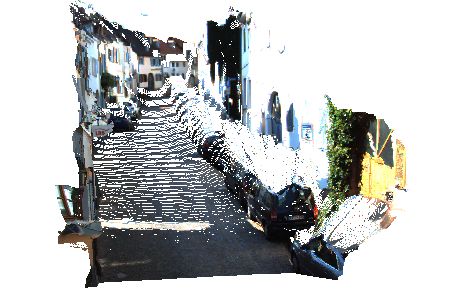} \\
\vspace{2mm}\includegraphics[width=0.22\linewidth, height=0.13\linewidth]{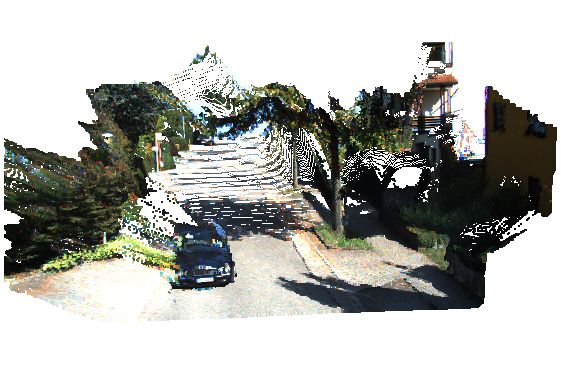} &
\includegraphics[width=0.22\linewidth, height=0.13\linewidth]{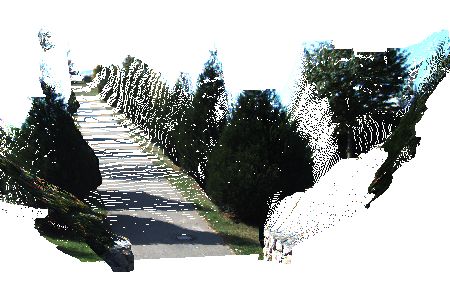} &
\includegraphics[width=0.22\linewidth, height=0.13\linewidth]{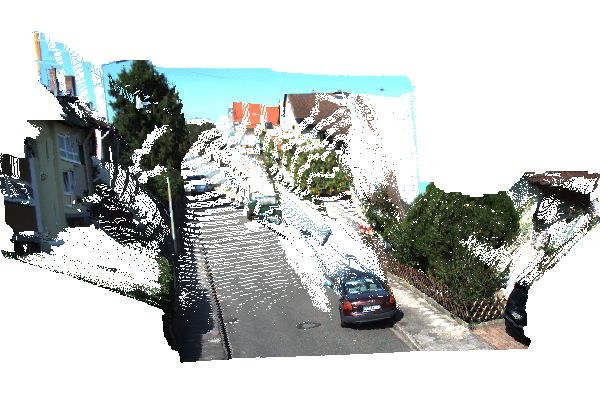} &
\includegraphics[width=0.22\linewidth, height=0.13\linewidth]{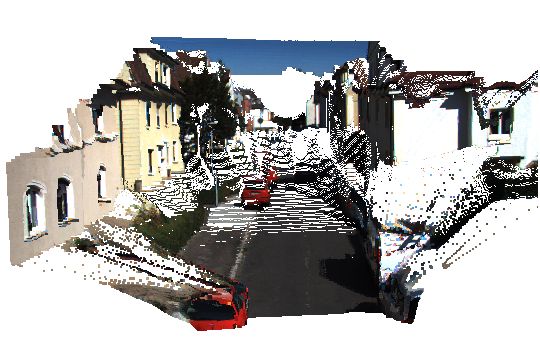} \\
\vspace{2mm}\includegraphics[width=0.22\linewidth, height=0.13\linewidth]{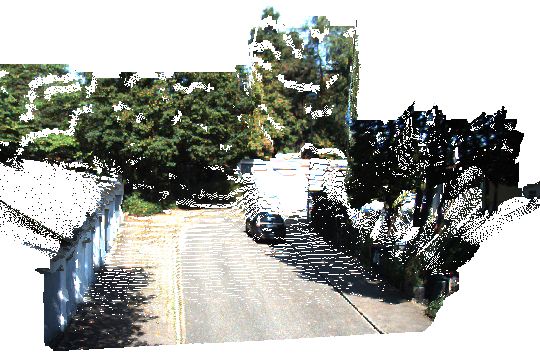} &
\includegraphics[width=0.22\linewidth, height=0.13\linewidth]{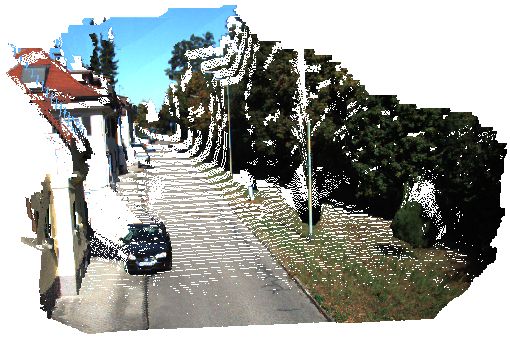} &
\includegraphics[width=0.22\linewidth, height=0.13\linewidth]{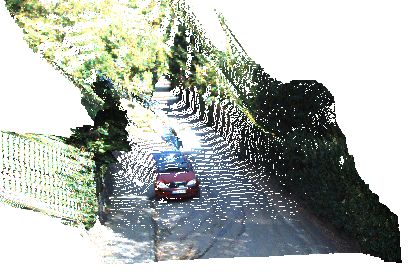} &
\includegraphics[width=0.22\linewidth, height=0.13\linewidth]{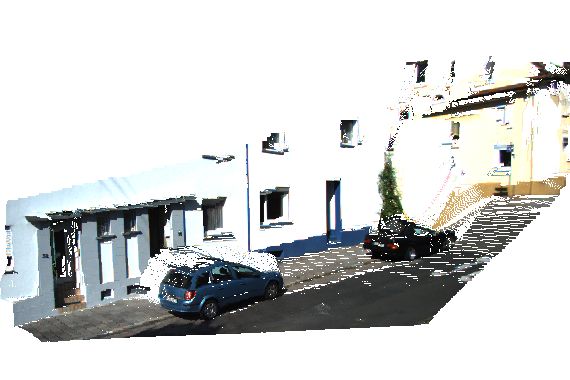}
\end{tabular}
 \small{\caption{Reconstructed 3D point clouds displayed from a different view point, obtained from the same pairs of images in the same order as in Figure~\ref{KITTI_res} and \ref{KITTI_comp}. The visually appealing synthetically rendered views give a much better idea of the quality of estimated depth.
\label{KITTI_3D}}}
\end{figure*}

The KITTI data set~\cite{Geiger2012CVPR} consists of $194$ pairs of training and $195$ pairs of test images, of the resolution approximately $1226 \times 370$, containing sparse disparity maps obtained by Velodyne laser scanner. All labelled points are in the bottom $2/3$rd of an image. Qualitative results are shown in Figure~\ref{KITTI_res}. The 3D point clouds obtained from the dense stereo correspondences are shown in Figure~\ref{KITTI_3D}. The Figure~\ref{KITTI_comp} shows the difference in the starting point between our classifier and standard feature-matching approaches, which after suitable regularization is also able to get close to the state-of-the-art.

Our classifier uses colour-based features, and thus was trained and evaluated on the pairs of images obtained by the pair of coloured cameras. The comparison on the evaluation server is done on the grey cameras instead. To compare our method to the state-of-the-art, we evaluated it on the colorized version of the grey images~\cite{Geiger2012CVPR}. The quality of the output images deteriorated slightly, partly also due to imprecise process of colourization. Quantitative comparisons can be found in the table~\ref{KITTI_q}.

\begin{table*}
\scriptsize
\begin{center}
\begin{tabular}{|c|c|c|c|c|}
\hline
Method & Result 3px & Result Occ 3px & Result 5px & Result Occ 5px\\
\hline
PCBP-SS~\cite{Yamaguchi13} & 3.40 \% &	4.72 \% & 2.18 \% & 3.15 \%\\
PCBP~\cite{Yamaguchi12a} & 4.04 \% & 5.37 \% & 2.64 \% & 3.64 \%\\
\hline
wSGM~\cite{Spangenberg13} & 4.97 \% & 6.18 \% & 3.25 \% & 4.11 \%\\
ATGV~\cite{Ranftl13} & 5.02 \% & 6.88 \% & 3.33 \% & 5.01 \%\\
iSGM~\cite{Hermann12} & 5.11 \% & 7.15 \% & 3.13 \% & 5.02 \%\\
ALTGV~\cite{Kuschk13} & 5.36 \% & 6.49 \% & 3.42 \% & 4.17 \%\\
ELAS~\cite{Geiger10} & 8.24 \% & 9.96 \% & 5.67 \% & 6.97 \%\\
OCV-BM~\cite{Bradski00} & 25.38 \% & 26.70 \% & 22.93 \% & 24.13 \%\\
GCut + Occ~\cite{Kolmogorov01computingvisual} & 33.49 \% & 34.73 \% & 27.39 \% & 28.62 \%\\
\hline
Our Result & 5.11 \% & 5.99 \% & 2.85 \% & 3.43 \%\\
\hline

\end{tabular}
\end{center}
\caption{\small Quantitative comparisons of our method with competing state-of-the-are approaches on the KITTI dataset in terms of ratio of pixels outside of the $3$ and $5$ pixel threshold, and in terms of average disparity error per pixel. All evaluations are done either on all pixels or except occluded pixels that are marked out. Approaches \cite{Yamaguchi13} and \cite{Yamaguchi12a} used all 21 pairs of frames in the videos. Our method gets comparable results to theirs, in particular for 5px threshold.}
\label{KITTI_q}
\end{table*}

\subsection{Sintel data set}
\begin{figure*}
\centering
\begin{tabular}{cccc}
\vspace{3.2mm}\includegraphics[width=0.23\linewidth, height=0.105\linewidth]{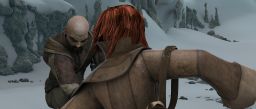} &
\includegraphics[width=0.23\linewidth, height=0.105\linewidth]{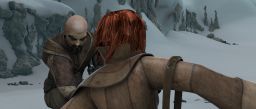} &
\includegraphics[width=0.23\linewidth, height=0.105\linewidth]{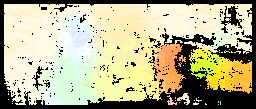} &
\includegraphics[width=0.23\linewidth, height=0.105\linewidth]{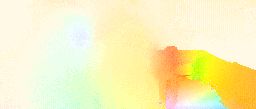} \\

\vspace{3.2mm}\includegraphics[width=0.23\linewidth, height=0.105\linewidth]{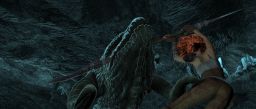} &
\includegraphics[width=0.23\linewidth, height=0.105\linewidth]{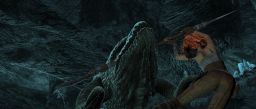} &
\includegraphics[width=0.23\linewidth, height=0.105\linewidth]{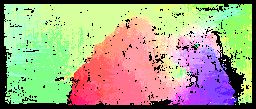} &
\includegraphics[width=0.23\linewidth, height=0.105\linewidth]{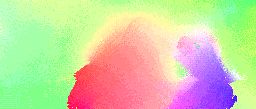} \\

\vspace{3.2mm}\includegraphics[width=0.23\linewidth, height=0.105\linewidth]{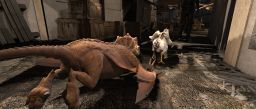} &
\includegraphics[width=0.23\linewidth, height=0.105\linewidth]{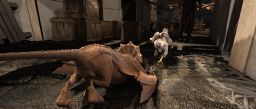} &
\includegraphics[width=0.23\linewidth, height=0.105\linewidth]{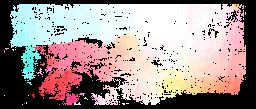} &
\includegraphics[width=0.23\linewidth, height=0.105\linewidth]{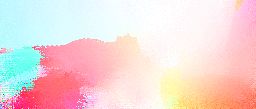} \\

\vspace{3.2mm}\includegraphics[width=0.23\linewidth, height=0.105\linewidth]{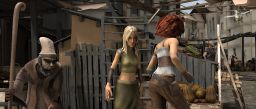} &
\includegraphics[width=0.23\linewidth, height=0.105\linewidth]{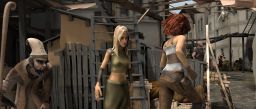} &
\includegraphics[width=0.23\linewidth, height=0.105\linewidth]{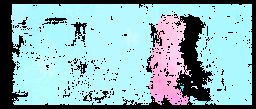} &
\includegraphics[width=0.23\linewidth, height=0.105\linewidth]{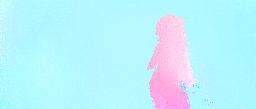} \\

\vspace{3.2mm}\includegraphics[width=0.23\linewidth, height=0.105\linewidth]{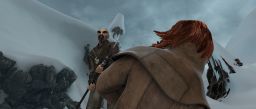} &
\includegraphics[width=0.23\linewidth, height=0.105\linewidth]{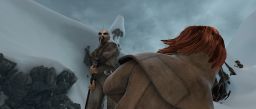} &
\includegraphics[width=0.23\linewidth, height=0.105\linewidth]{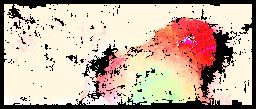} &
\includegraphics[width=0.23\linewidth, height=0.105\linewidth]{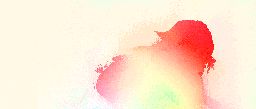} \\

\vspace{3.2mm}\includegraphics[width=0.23\linewidth, height=0.105\linewidth]{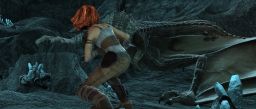} &
\includegraphics[width=0.23\linewidth, height=0.105\linewidth]{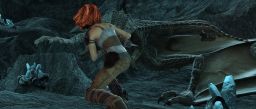} &
\includegraphics[width=0.23\linewidth, height=0.105\linewidth]{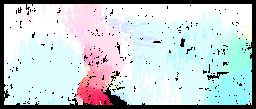} &
\includegraphics[width=0.23\linewidth, height=0.105\linewidth]{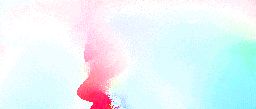} \\

\vspace{3.2mm}\includegraphics[width=0.23\linewidth, height=0.105\linewidth]{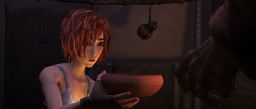} &
\includegraphics[width=0.23\linewidth, height=0.105\linewidth]{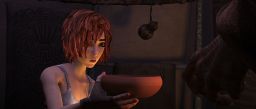} &
\includegraphics[width=0.23\linewidth, height=0.105\linewidth]{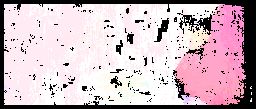} &
\includegraphics[width=0.23\linewidth, height=0.105\linewidth]{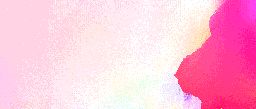} \\

\vspace{3.2mm}\includegraphics[width=0.23\linewidth, height=0.105\linewidth]{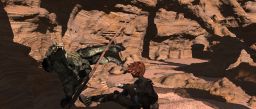} &
\includegraphics[width=0.23\linewidth, height=0.105\linewidth]{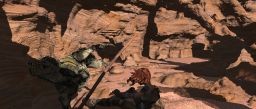} &
\includegraphics[width=0.23\linewidth, height=0.105\linewidth]{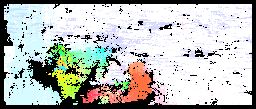} &
\includegraphics[width=0.23\linewidth, height=0.105\linewidth]{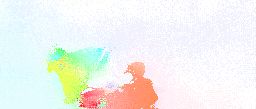} \\

Reference image & Matched image & Unary result & Regularized result\\
\end{tabular}

\vspace{0.4cm}
\raggedright
\parbox{14.8cm}{\vspace{-1.5cm}\caption{Qualitative results on Sintel dataset. For visualization we hand-picked pairs of images with larger range of displacement. The amplitude of the flow is normalized by the maximum flow in each individual image and colour-coded using the flow chart displayed on the right. Unary result shown is after the inverse matching validation step to remove the dependency of colour coding to noise. Black pixels correspond to pixels, that did not pass the validation test. The results suggest, our approach could be used as an initial step of the coarse-to-fine strategy for optical flow problem.
\label{sintel_clean}}}
\mbox{ } \includegraphics[width=0.12\linewidth, height=0.12\linewidth]{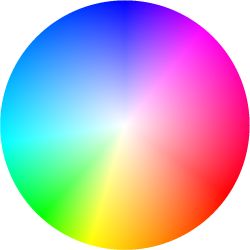}
\end{figure*}

The Sintel dataset (clean version) consists of the short clips from the rendered open source movie Sintel. It contains $23$ training and $12$ test clips, each one from $20$ to $50$ frames long. Together there are $1041$ training and $552$ test pairs of images of the resolution $1024 \times 436$. Most pixel flows are in the range of $(-160, 160) \times (-160, 160)$. Due to memory limitations, the evaluation of our classifier is not feasible for such a large range of flows. Thus, we trained and evaluated our method on the sub-sampled images (and thus also flows) by a factor of $4\times4$. Qualitative results are shown in the figure~\ref{sintel_clean}. We could not compare quantitatively to other methods for sub-sampled images, because the ground truth labelling is not available. Our classifier showed the potential to be used as a preprocessing step for another optical flow method; either as an initialization for its inference or as a way to restrict an allowed range of possible flows for high resolution images in a coarse-to-fine fashion.

\subsection{TimeLapse data set}
\begin{figure*}
\centering
\begin{tabular}{ccccccc}
\vspace{3.2mm}\begin{sideways}\hspace{4mm}{\scriptsize{Frame 1}}\end{sideways} &
\includegraphics[width=0.14\linewidth]{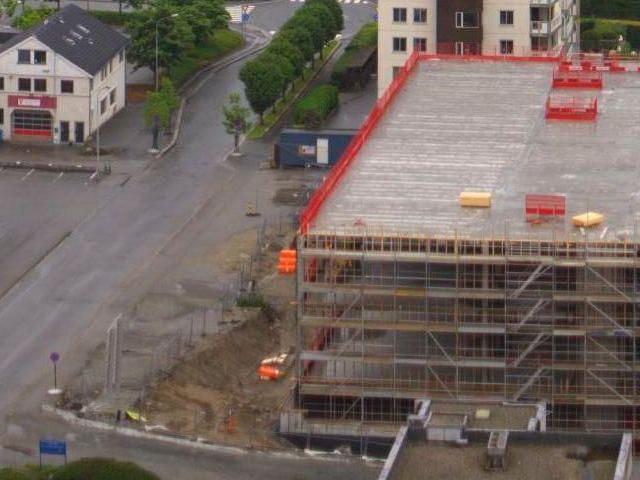} &
\includegraphics[width=0.14\linewidth]{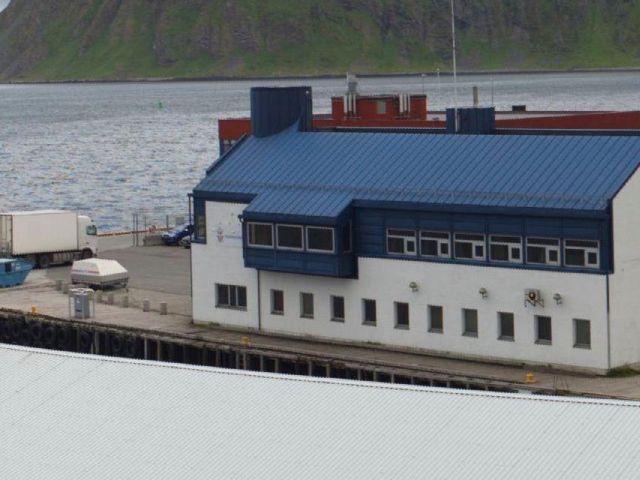} &
\includegraphics[width=0.14\linewidth]{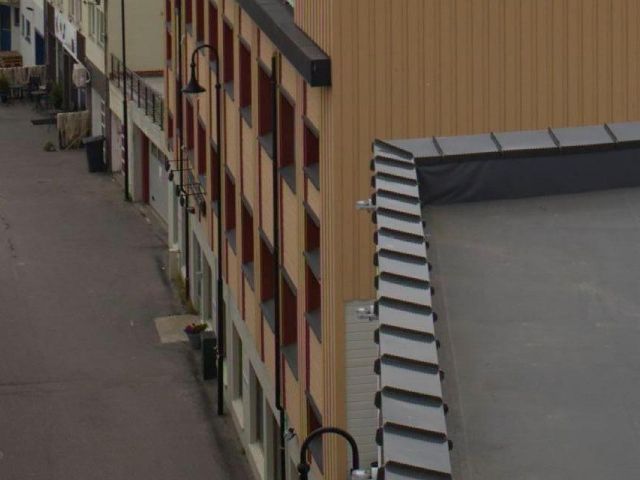} &
\includegraphics[width=0.14\linewidth]{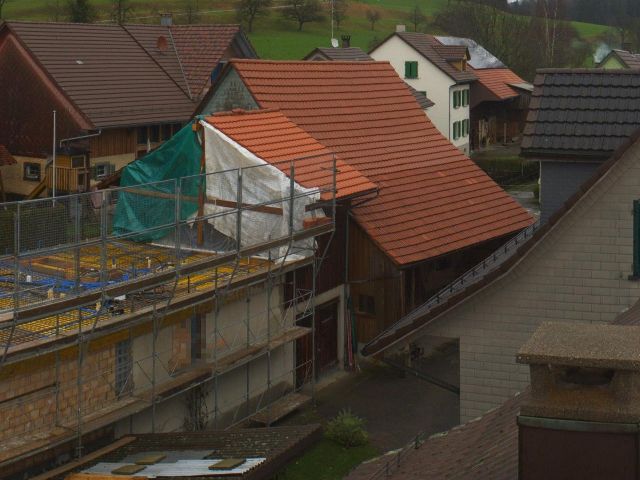} &
\includegraphics[width=0.14\linewidth]{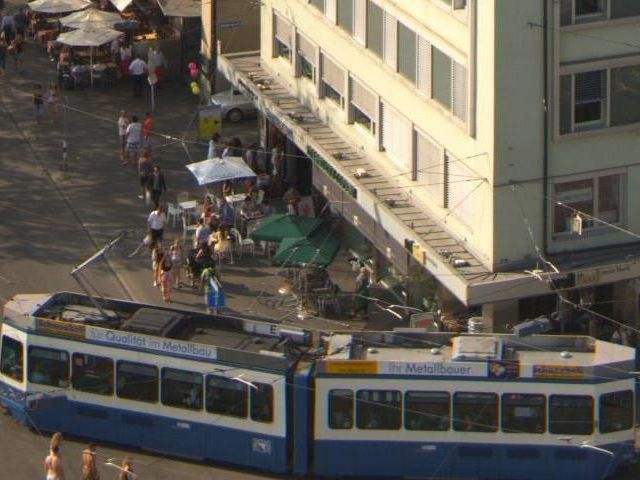} &
\includegraphics[width=0.14\linewidth]{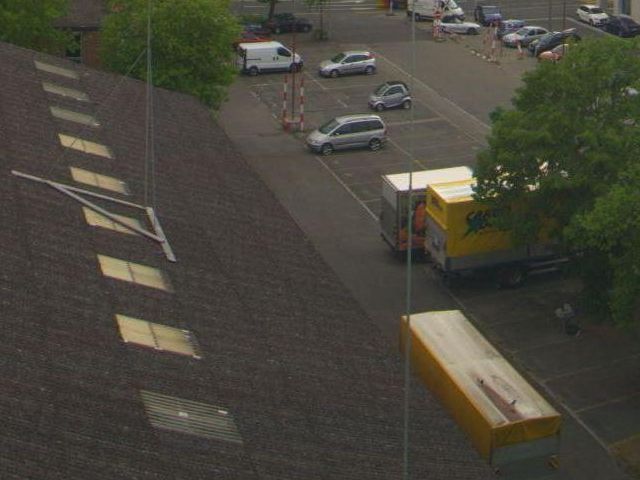} \\
\vspace{3.2mm}\begin{sideways}\hspace{4mm}{\scriptsize {Frame 2}}\end{sideways} &
\includegraphics[width=0.14\linewidth]{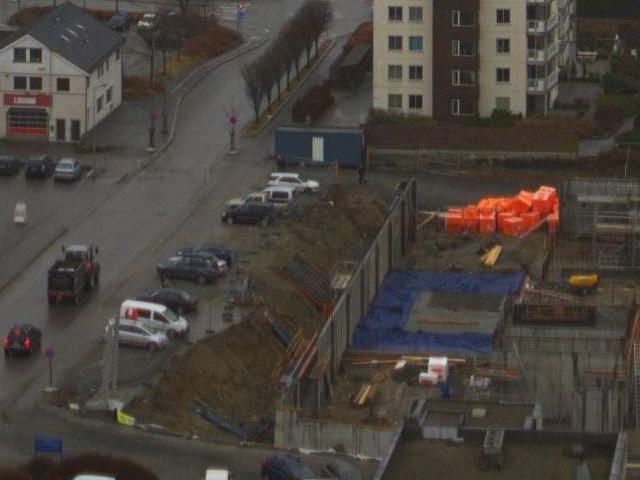} &
\includegraphics[width=0.14\linewidth]{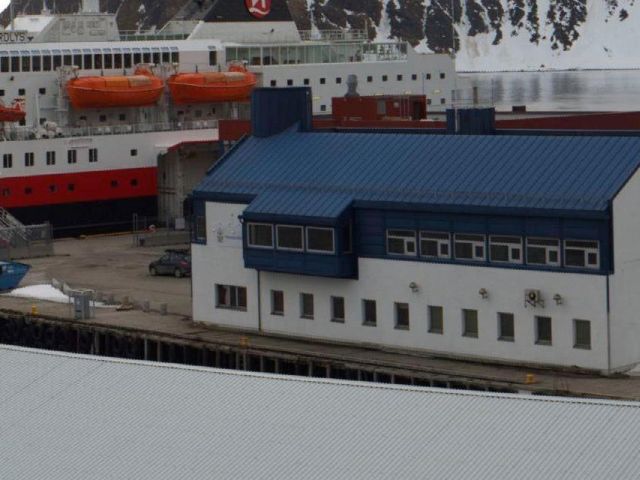} &
\includegraphics[width=0.14\linewidth]{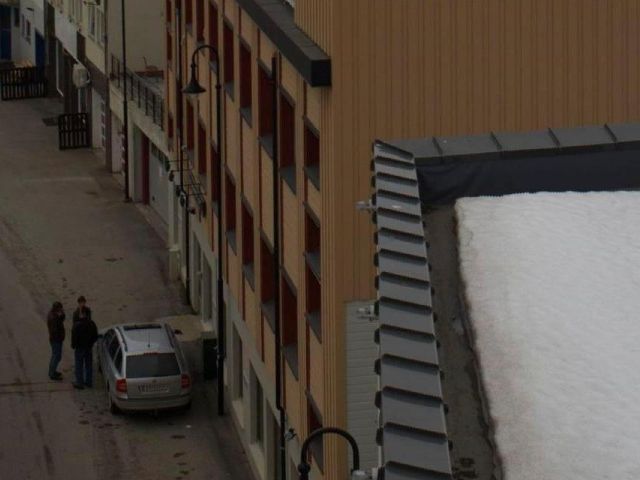} &
\includegraphics[width=0.14\linewidth]{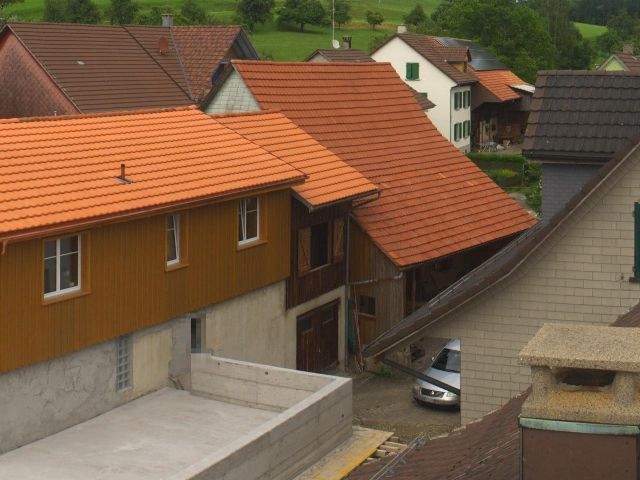} &
\includegraphics[width=0.14\linewidth]{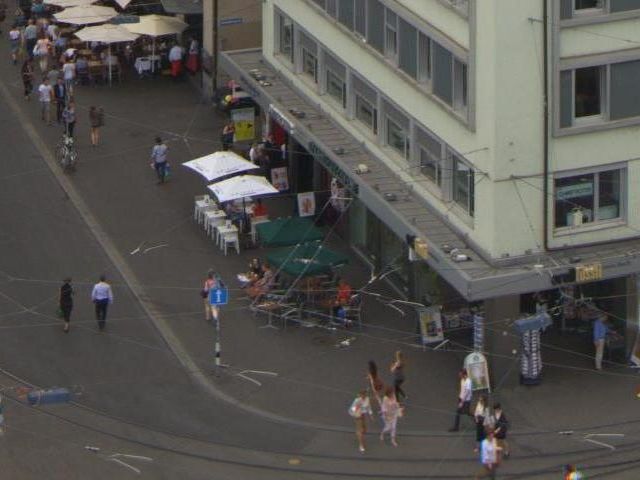} &
\includegraphics[width=0.14\linewidth]{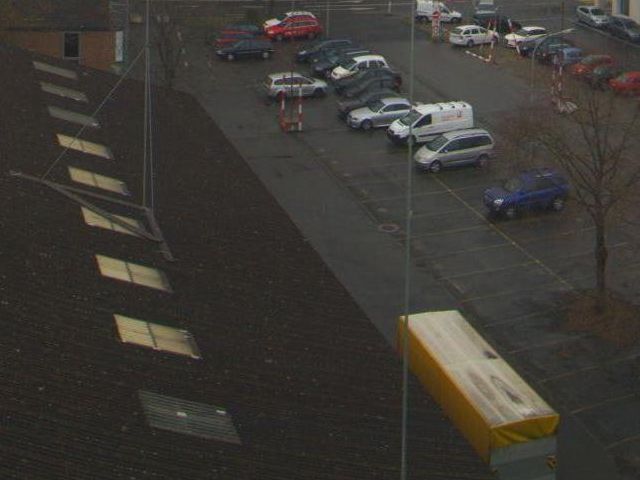} \\
\vspace{3.2mm}\begin{sideways}\hspace{3mm}{\scriptsize{Our Result}}\end{sideways} &
\includegraphics[width=0.14\linewidth]{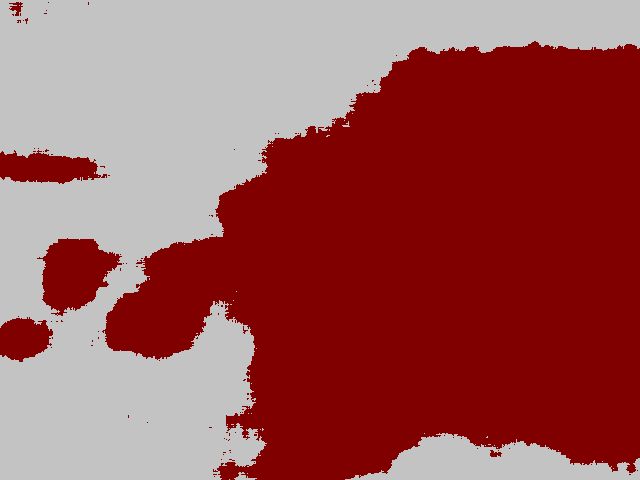} &
\includegraphics[width=0.14\linewidth]{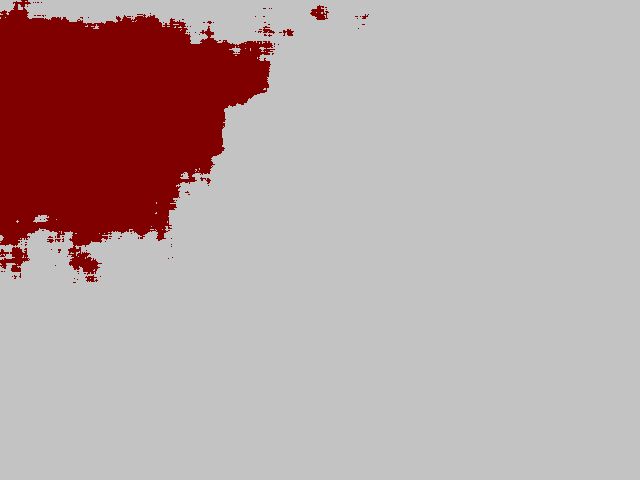} &
\includegraphics[width=0.14\linewidth]{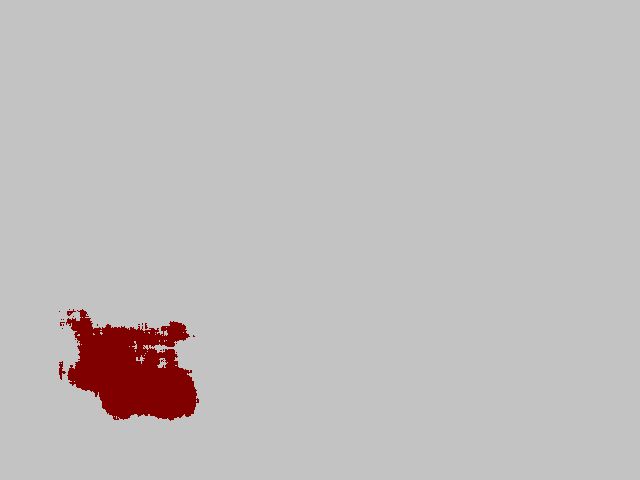} &
\includegraphics[width=0.14\linewidth]{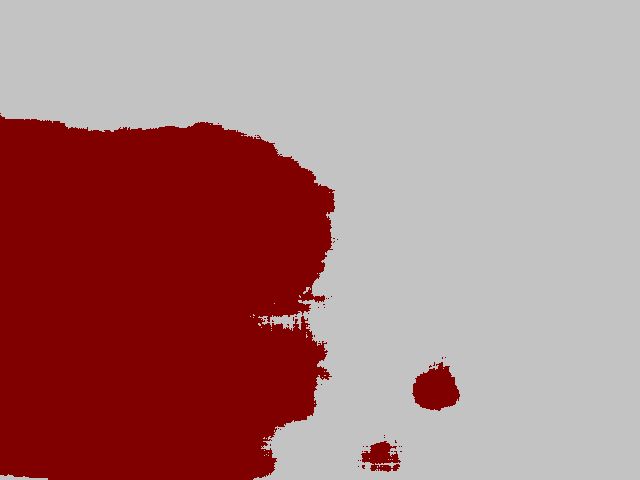} &
\includegraphics[width=0.14\linewidth]{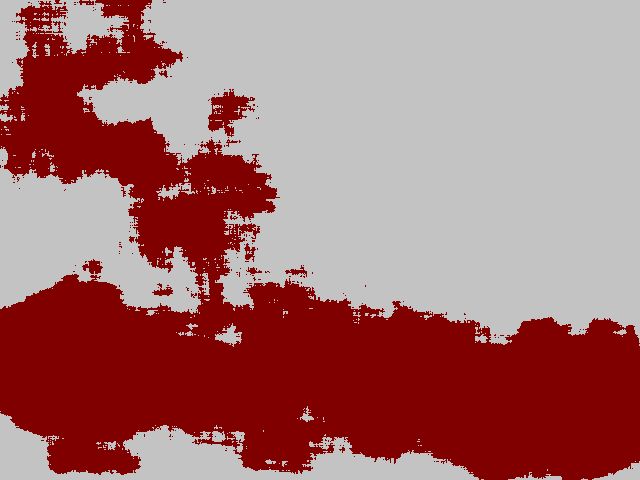} &
\includegraphics[width=0.14\linewidth]{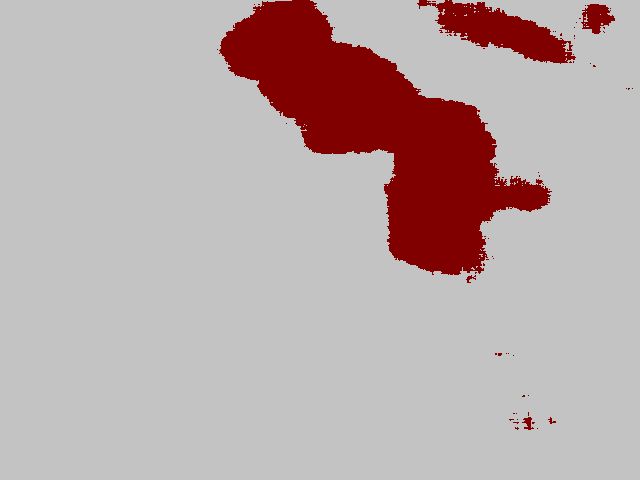} \\
\vspace{3.2mm}\begin{sideways}\hspace{1mm}{\scriptsize{Ground Truth}}\end{sideways} &
\includegraphics[width=0.14\linewidth]{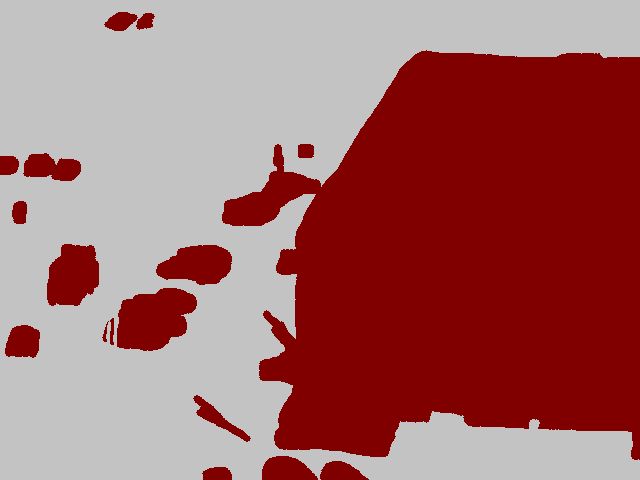} &
\includegraphics[width=0.14\linewidth]{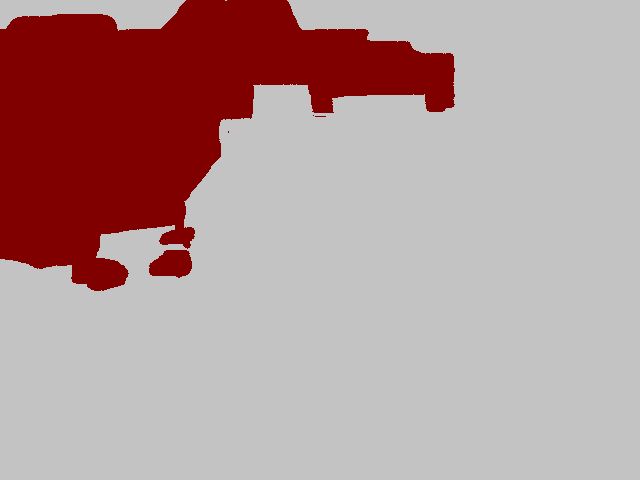} &
\includegraphics[width=0.14\linewidth]{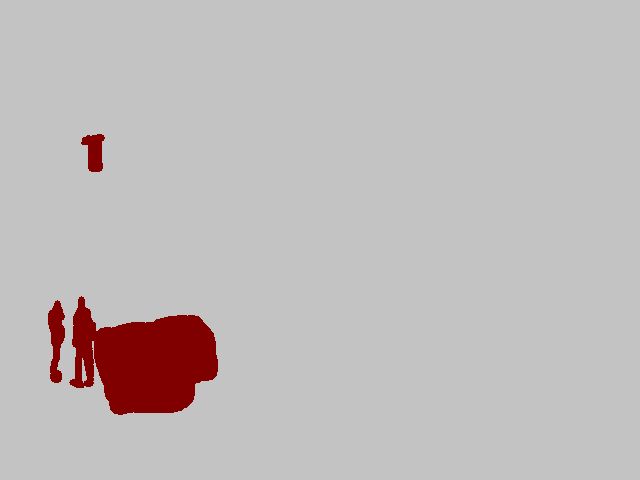} &
\includegraphics[width=0.14\linewidth]{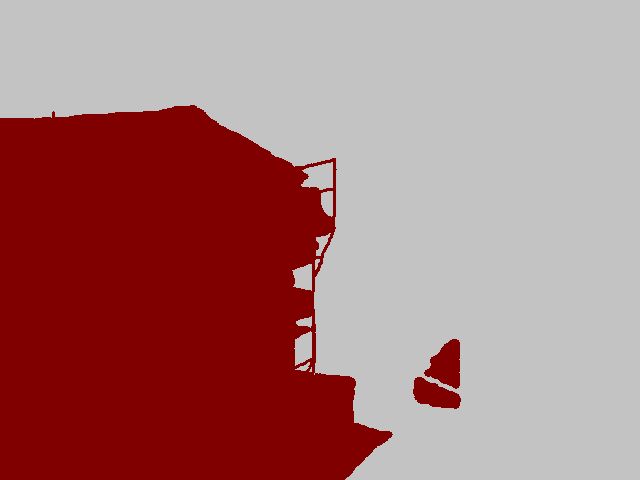} &
\includegraphics[width=0.14\linewidth]{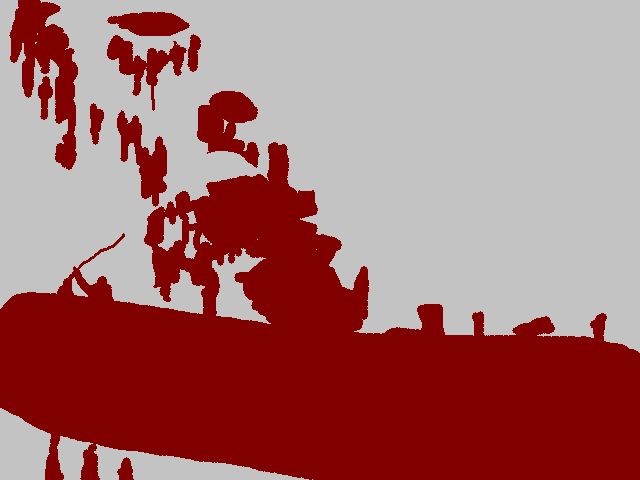} &
\includegraphics[width=0.14\linewidth]{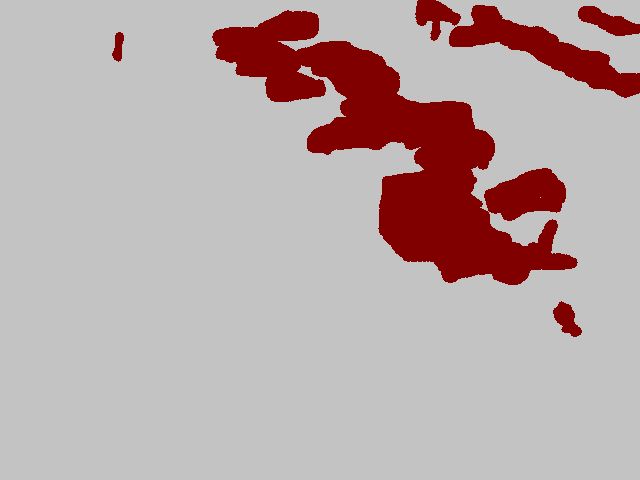}
\end{tabular}
 \small{\caption{Qualitative results on the TimeLapse dataset. Our classifier managed to distinguih between structural changes and seasonal changes. The appearance of trees in 1st and 6th image have changed dramatically, however, our classifier did not label it as a change. Similarly the snow in the background of 2rd and on the roof of 3th image was not labelled as change. On the other hand, structural changes, such as new building built, have been consistently recognized as a change.
\label{time_res}}}
\end{figure*}

In this paper we also introduce a new TimeLapse temporal change detection dataset. It contains 134 pairs of training and 50 pairs of test images and annotated pixel-wise ground truth indicating the structural change in the scene, such as new buildings, tents, vehicles or construction sites. Lighting or seasonal changes (snow, leaves on the tree) are not labelled as change. The dataset will be made available on publication. Qualitative results of our classifier are shown in Figure~\ref{time_res}. Our classifier successfully managed to distinguish between seasonal changes and structural changes in a scene. Quantitatively, $93.3\%$ of pixels were correctly labelled, average recall per class was $86.9\%$, average precision $87.9\%$.

\section{Conclusions}
Our contextual matching classifier showed the potential to be a promising direction of further research in the this area - the application of discriminative training for this task. We validated our approach on three challenging problems - stereo matching, optical flow and pixel-wise temporal change detection. In our further work we would like to test various different design choices to maximize the quantitative performance of our classifier. It might give us some insights, how to design the discriminatively trained classifier, that would generalize across different data sets.

\bibliographystyle{splncs}
\bibliography{match_cvpr15}

\end{document}